\documentclass[final]{cvpr}

\usepackage{times}
\usepackage{epsfig}
\usepackage{graphicx}
\usepackage{amsmath}
\usepackage{amssymb}
\usepackage{subfig}
\usepackage{diagbox}
\usepackage{enumitem}
\usepackage{rotating} 
\usepackage{cases}
\usepackage{bm}
\usepackage{color}
\usepackage{colortbl}
\usepackage{booktabs}
\usepackage{xspace}
\definecolor{mygray}{gray}{.95}

\newcommand{\M}[1]{\mathbf{#1}}

\definecolor{wincolor}{rgb}{0.95, 0.2, 0.2}

\newcommand{\dataset}[1]{{\fontfamily{cmtt}\selectfont #1} }
\newcommand{\Sun}{\dataset{SUN360}}

\def\gb{Gr{\"o}bner basis\xspace}

\usepackage[pagebackref=true,breaklinks=true,colorlinks,bookmarks=false]{hyperref}

\def\cvprPaperID{6981} 
\def\confYear{CVPR 2021}

\begin{document}

\title{Minimal Solutions for Panoramic Stitching Given Gravity Prior}

\author{Yaqing Ding$^{1}$, Daniel Barath$^{2,3}$, Zuzana Kukelova$^{2}$\\
$^1$ School of Computer Science and Engineering, 
Nanjing University of Science and Technology\\
$^2$ Visual Recognition Group, Faculty of Electrical Engineering, 
Czech Technical University in Prague \\
$^3$ Machine Perception Research Laboratory, 
SZTAKI, Budapest \\
{\tt\small dingyaqing@njust.edu.cn}
}

\maketitle

\begin{abstract}
    When capturing panoramas, people tend to align their cameras with the vertical axis, i.e., the direction of gravity.  
    Moreover, modern devices, such as smartphones and tablets, are equipped  with an IMU (Inertial Measurement Unit) that can measure the gravity vector accurately.
    Using this prior, the $y$-axes of the cameras can be aligned or assumed to be already aligned, reducing their relative orientation to 1-DOF (degree of freedom).
    Exploiting this assumption, we propose new minimal solutions to panoramic image stitching of images taken by cameras with coinciding optical centers, i.e., undergoing pure rotation.
    We consider four practical camera configurations, assuming unknown fixed or varying focal length with or without radial distortion. 
    The solvers are tested both on synthetic scenes and on more than 500k real image pairs from the Sun360 dataset and from scenes captured by us using two smartphones equipped with IMUs.
    It is shown, that they outperform the state-of-the-art both in terms of accuracy and processing time. 
\end{abstract}

\section{Introduction}\label{sec:intro}

Panoramic image stitching is a fundamental problem in computer vision. 
When solving this problem, we are given a sequence of images taken from a single point in space with a camera rotating around some 3D axis.  
The objective is to map the images into a common reference frame and to create a larger image composed of the captured ones, thus, covering a much wider field-of-view than each individual image. 
In other words, the goal is to estimate the unknown relative rotation and inner calibration parameters of cameras with coinciding optical centers, \ie, cameras undergoing a pure rotational motion.
In the computer vision community, this problem is often considered to be solved with a number of existing solutions~\cite{hartley2003multiple,brown2007minimal,fitzgibbon2001simultaneous,jin2008three,byrod2009minimal,kukelova2015radial}.
However, in this paper, we will show that the existing solutions do not exploit all available information which can be easily obtained from recent devices. This information can be used to simplify both the problem formulation and solution.

\begin{figure}[t]
    \centering
  	    \subfloat{ \includegraphics[trim={0mm 5mm 0mm 0mm},clip,width=1.0\columnwidth]{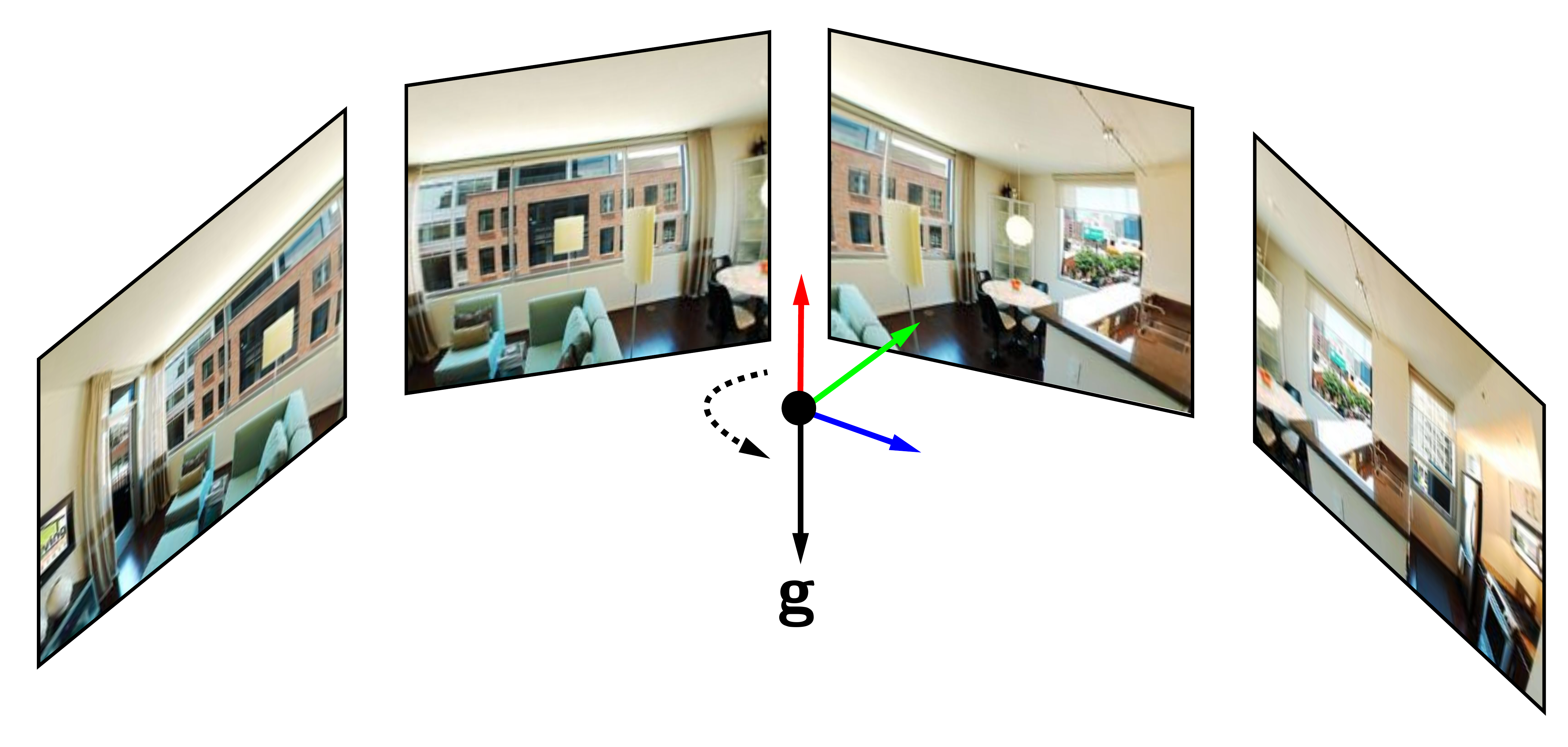}}\\[-2mm]
    \caption{Panorama stitching with known gravity vector $\M g$.}
    \label{fig:teaser}
\end{figure}

A typical panorama stitching pipeline consists of the following two major steps.
\begin{enumerate}
    \item \textit{Pair-wise computation}: 
    In the first step, image features are obtained and tentative correspondences are matched between all image pairs.
    From these correspondences, inner and outer calibration parameters of pairs of cameras are estimated robustly and the correspondences consistent with the calibration are selected. The robust estimation is usually based on solving the stitching problem from a minimal number of input correspondences, \ie, solving the problem using a minimal solver, in a RANSAC framework~\cite{brown2007minimal}.
    \item \textit{Bundle adjustment (BA)}: The pair-wise estimates of inner and outer calibration parameters are refined jointly over all images using a non-linear optimization. 
\end{enumerate}

\begin{figure*}[t]
    \centering
        \subfloat[Full panoramic image]{
        \begin{tabular}[b]{c}  
    	    \scalebox{-1}[1]{\includegraphics[width=0.47\textwidth]{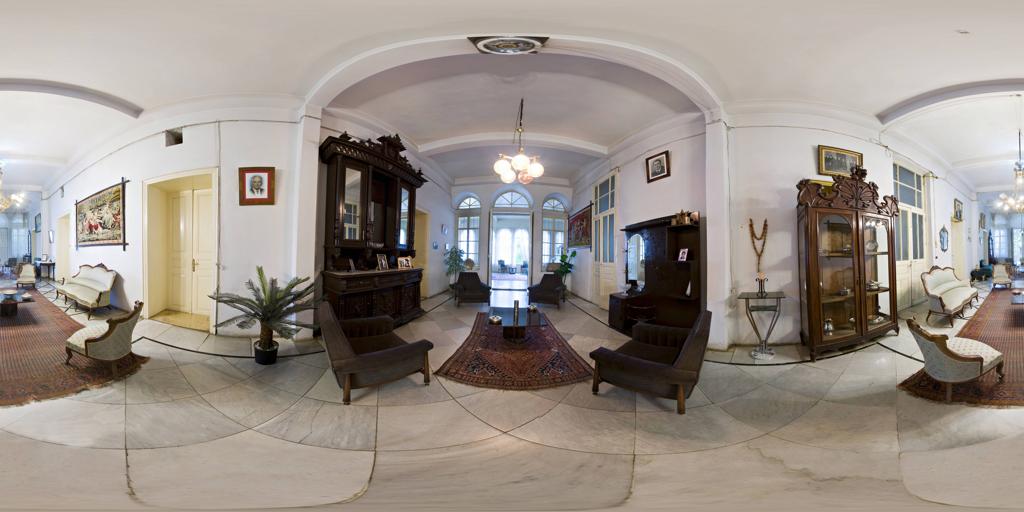}}\\
    	    \scalebox{-1}[1]{\includegraphics[width=0.47\textwidth]{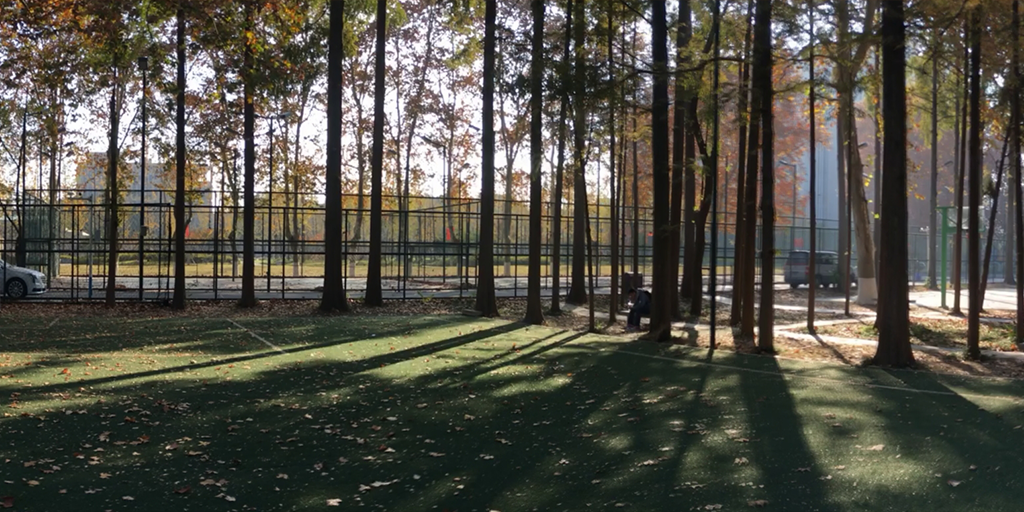}}
	     \end{tabular}}
	    \subfloat[Image pair]{\begin{tabular}[b]{c}  
	        \scalebox{-1}[1]{\includegraphics[width=0.1515\textwidth]{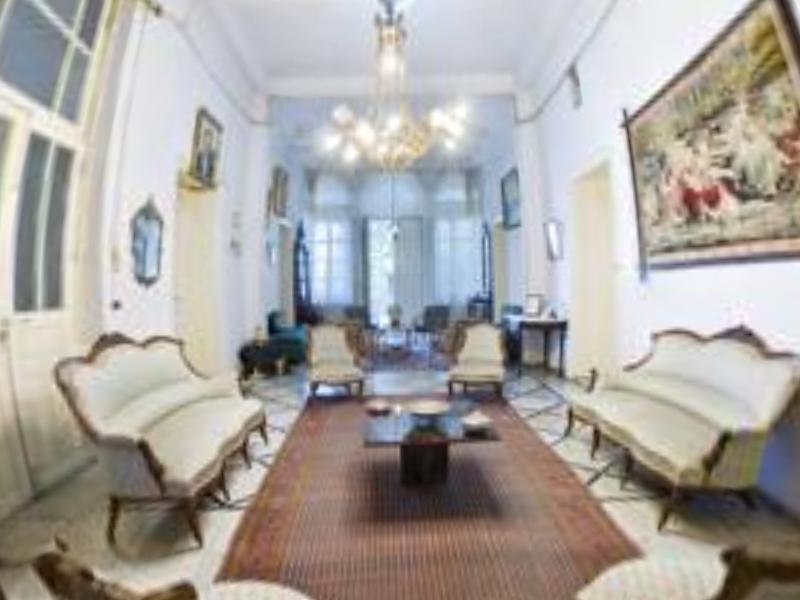}}\\        \scalebox{-1}[1]{\includegraphics[width=0.1515\textwidth]{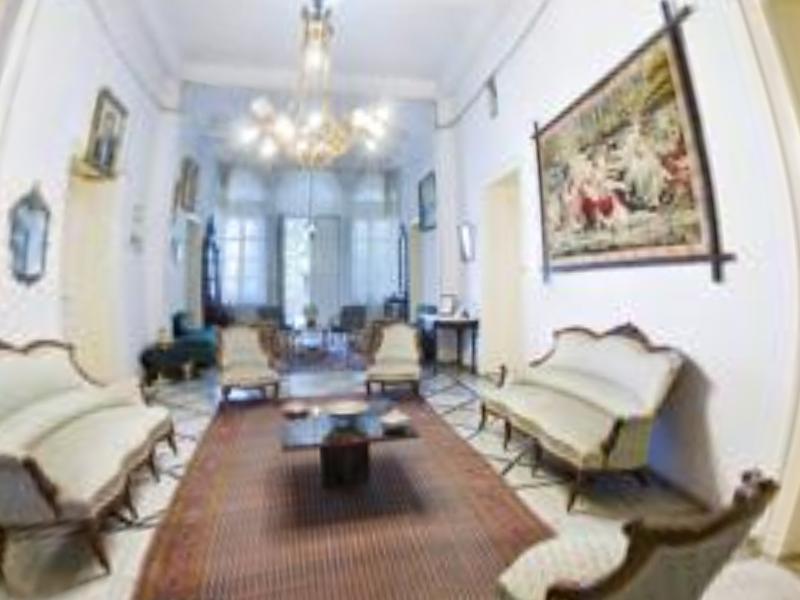}}\\
	        \scalebox{-1}[1]{\includegraphics[trim={0mm 0mm 112mm 0mm},clip,width=0.1515\textwidth]{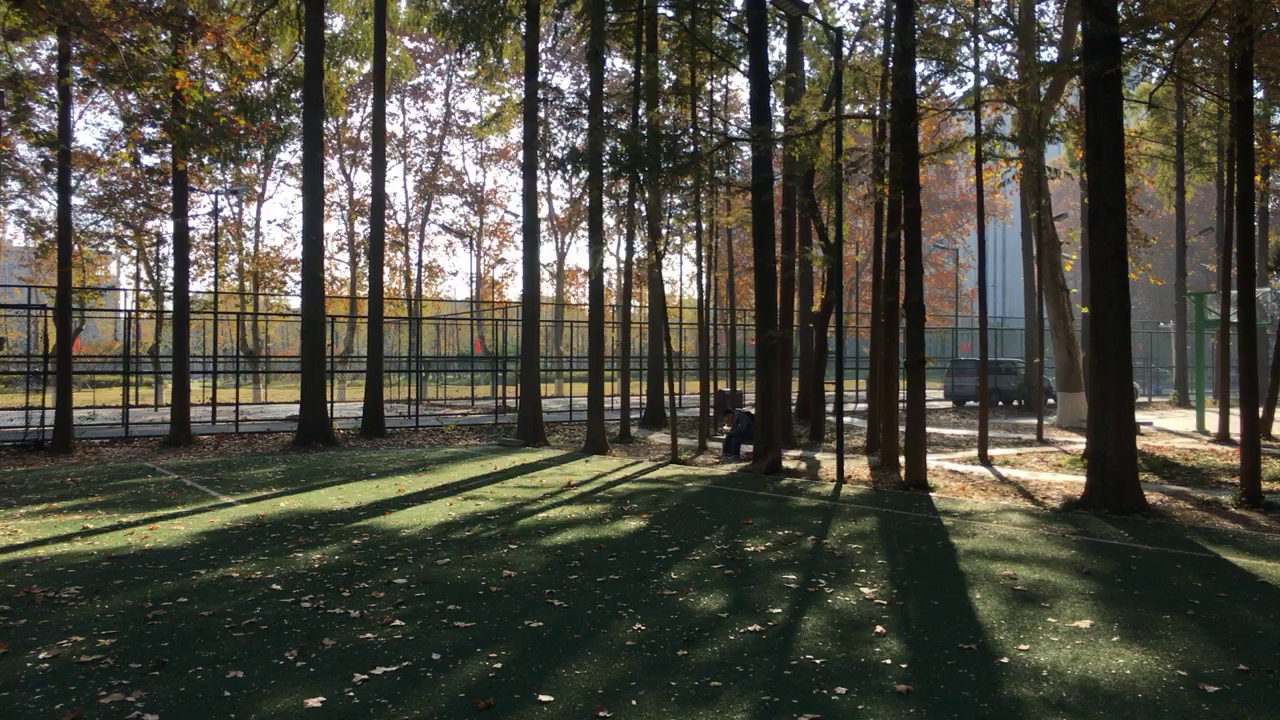}}\\        \scalebox{-1}[1]{\includegraphics[trim={0mm 0mm 112mm 0mm},clip,width=0.1515\textwidth]{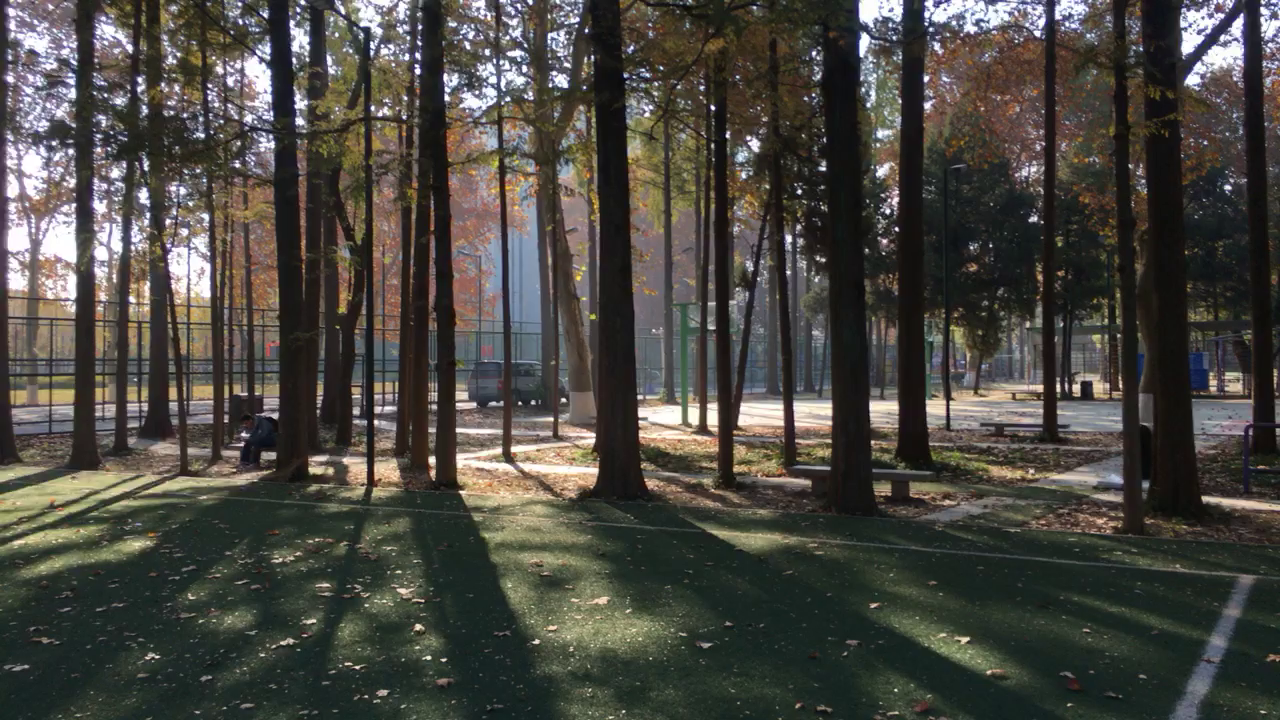}}
	     \end{tabular}}
	    \subfloat[Stitching]{
        \begin{tabular}[b]{l} 
	        \scalebox{-1}[1]{\includegraphics[width=0.36\textwidth]{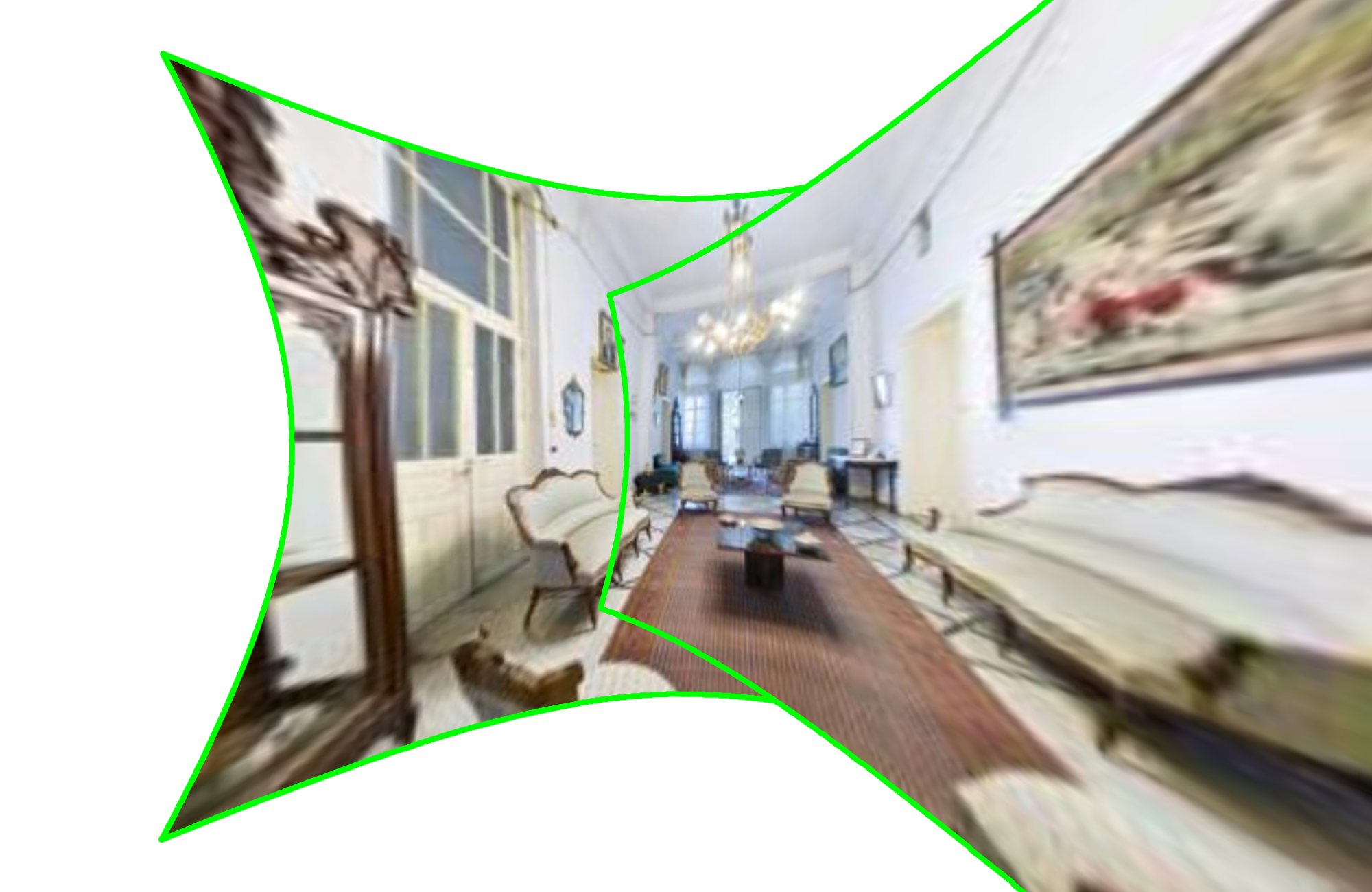}}\\
	        \scalebox{-1}[1]{\includegraphics[trim={115mm 93mm 197mm 78mm},clip,width=0.32\textwidth]{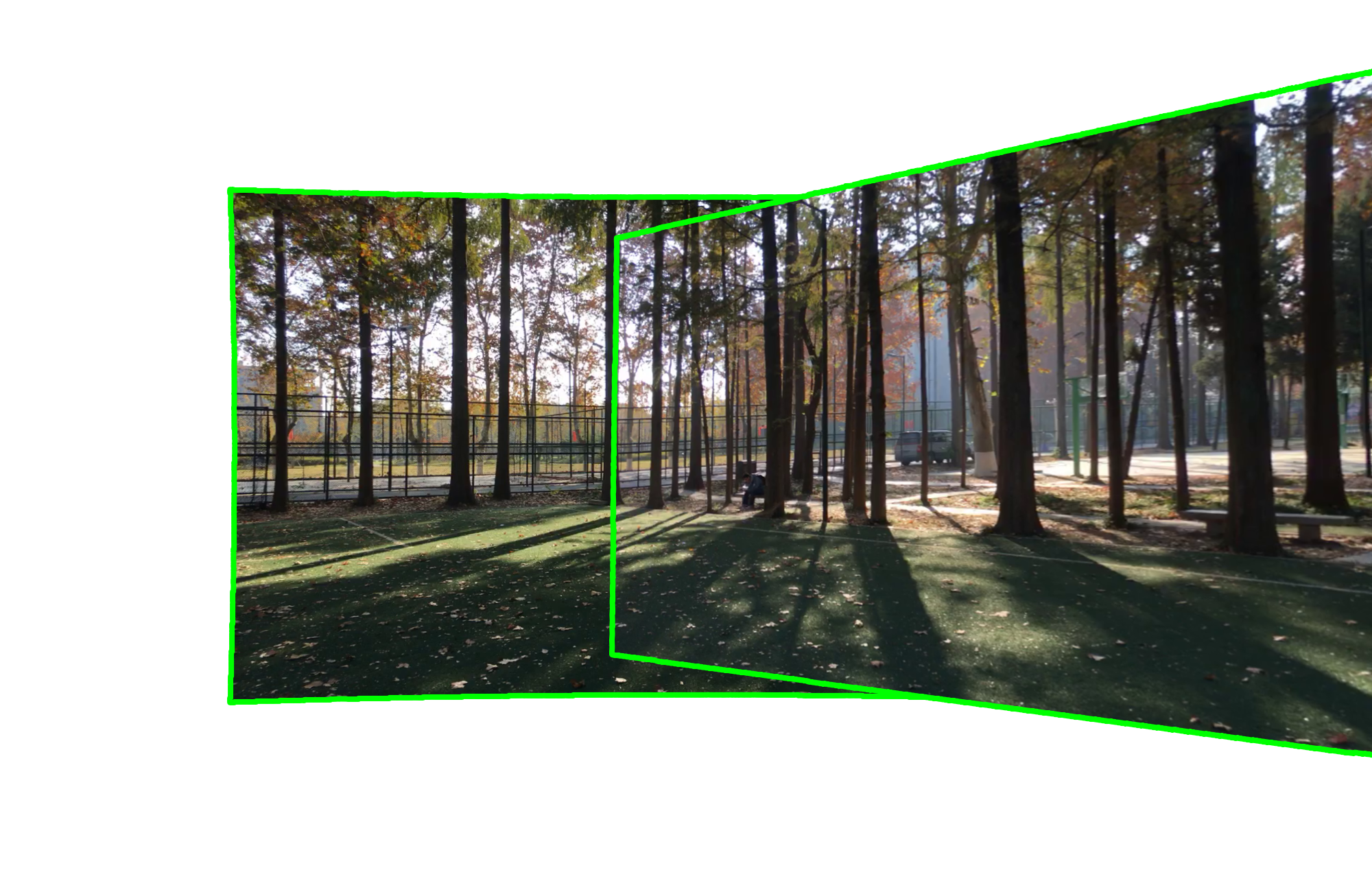}}
	     \end{tabular}}
    \caption{ Example scenes and stitching results from the \Sun (top) and the created Smartphone (bottom) datasets. }
    \label{fig:real_experiments}
\end{figure*}

Several minimal solvers exists which can be used in the pair-wise step of the panoramic stitching pipeline.
One of the basic well-known solvers for panoramic stitching assumes fully calibrated pinhole cameras and estimates the unknown rotation from two point correspondences~\cite{horn1987closed}. 
However, the camera intrinsic parameters may not be known in practice, which is often the case in real-world. Solutions to different camera configurations can be divided into two main categories: the unknown focal length case for narrow field-of-view cameras, and the unknown 
radial distortion case for wide-angle cameras. For the unknown focal length case, one of the most commonly used solvers is the normalized 4 point linear solution for homography estimation~\cite{hartley2003multiple}. In~\cite{brown2007minimal}, the authors demonstrated that 2 and 3 points are sufficient for obtaining the homography induced by a rotation with 1 and 2 unknown focal length parameters. 

In practice, almost all cameras exhibit some amount of lens distortion. 
Moreover, wide-angle lenses with significant radial lens distortion are now commonly used in smart devices. 
These wide-angle cameras introduce large distortions that can not be ignored.
In~\cite{fitzgibbon2001simultaneous,jin2008three}, it was shown that modeling lens distortion inside the minimal solver is critical for obtaining high-quality homography estimates.  Without modeling lens distortion in the pair-wise step, not only the initial pose and calibration estimates are less accurate but the set of inliers used for the non-linear optimization does not cover the whole image and misses especially highly distorted points from image borders. Without such points, the non-linear optimization in the BA stage may not be  able to correctly estimate the lens distortion and to produce good stitching results.
To address the case where there is lens distortion in the images, Fitzgibbon~\cite{fitzgibbon2001simultaneous} used the single parameter division model to simultaneously estimate the homography and radial distortion using 5 point correspondences. With this division model, the minimal 3 point solver was proposed for panoramic stitching with equal and unknown radial distortion~\cite{jin2008three,byrod2009minimal}. In~\cite{kukelova2015radial}, two solvers were proposed for estimating homography between two cameras with different radial distortions.
 
All the mentioned minimal solvers estimate the full 3-DOF rotation. This, especially for cameras with radial distortion, makes the resulting system of polynomial equations complicated.
However, panoramas are usually captured with cameras or mobile phones aligned with the direction of gravity. Moreover, recent devices usually are equipped with an IMU sensor that can measure the gravity vector accurately. Using the gravity prior, the $y$-axes of the cameras can be aligned, reducing their relative orientation to 1-DOF.
This prior not only simplifies the geometry and polynomial systems that have to be solved but, also, reduces that number of correspondences needed for the estimation. This is extremely important since the processing time of RANSAC-like robust estimation depends \textit{exponentially} on the sample size.
The gravity prior was used to simplify different minimal relative~\cite{fraundorfer2010minimal,naroditsky2012two,sweeney2014solving,2014Relative,saurer2017homography,ding2019efficient,Ding_2020_CVPR} and absolute pose solvers~\cite{kukelova2010closed,2015Efficient,albl2016rolling}. Surprisingly, it was not considered in the panoramic stitching solvers.

In this paper, we present the first minimal solutions to panoramic image stitching of images taken by two cameras with coinciding optical centers, \ie, undergoing pure rotation, under the assumption of known gravity direction. 
We consider four practical camera configurations:
\begin{enumerate}[label=(\roman*)]
\item ${\bf H1f}$ - {\bf Equal focal length}: The images are captured by a camera with fixed unknown focal length and known or negligible radial distortion (narrow field-of-view). 
\item ${\bf H2f_1f_2}$ - {\bf Varying focal lengths}: The images are captured by cameras with different focal lengths, \eg, by a zoom camera, with negligible or known distortion. 
\item ${\bf H2\lambda}$ - {\bf Equal focal length and radial distortion}: The images are captured by a camera with fixed unknown focal length and radial distortion. 
\item ${\bf H3\lambda_1\lambda_2}$ - {\bf Varying focal lengths and radial distortion}:  The most general case, where the images are captured by cameras with different focal lengths and distortions, \eg, using a wide-angle camera that was zooming during the image capture.
\end{enumerate}

Experiments both on synthetic data and more than $500$k real image pairs demonstrate that our new solvers are superior to the state-of-the-art methods both in terms of accuracy and computational efficiency, \ie, the run-time.

\section{Problem Statement}\label{section3}

Let us assume that we are given a set of 3D points $\left\{\M{X}_i\right\}$ observed by two cameras undergoing a pure rotational motion. Let $\M p^d_{1i} = [u^d_{1i},v^d_{1i},1]^\top$ and $\M p^d_{2i} = [u^d_{2i},v^d_{2i},1]^\top$ be the corresponding  measured distorted image projections of $\M X_i$ in these two cameras in their homogeneous form. 
Undistorted image points $\M p^u_{1i} =  u(\M p^d_{1i},\M \bm{\lambda_1}) $ and $\M p^u_{2i} =  u(\M p^d_{2i},\M \bm{\lambda_2}) $, undistorted with some undistoriton function $u(\cdot,\bm{\lambda})$, are related by
\begin{equation}
d_{2i} \M K{_2^{-1}} \M p^u_{2i} = d_{1i} \M R \M K{_1^{-1}} \M p^u_{1i} , \label{eq:basic_relation}
\end{equation}
where $d_{1i},d_{2i}$ are the projective depths, $\M K_1, \M K_2$ are the intrinsic camera matrices, and $\M R\in \text{SO}(3)$ is the unknown relative rotation between the cameras. 

We use the one-parameter division model to parameterize radial undistortion function $u$. 
This model suits especially for minimal solvers since it is able to express a wide range of distortions with a single parameter and usually results in simpler equations compared with other distortion models.
This model was used, \eg, by \cite{fitzgibbon2001simultaneous} for the simultaneous estimation of two-view geometry and lens distortion, for the estimation of radial distortion homography~\cite{kukelova2015radial}, and, also, for panorama stitching with radial distortion~\cite{byrod2009minimal,jin2008three}. 

In the one parameter division model, the undistortion function $u$, that is undistorting distorted image point with homogeneous coordinates $\M p^d = [u^d,v^d,1]$ using the undistortion parameter $\lambda$ has the following form 
\begin{equation}
\M p^u = u(\M p^d,\M \lambda) = [u_{d}, v_{d}, 1+\lambda (u_{d}^{2}+v_{d}^{2})]^\top.
\label{eq:dic}
\end{equation}

In this paper, we assume that the gravity direction is known. This is a reasonable assumption since panoramas are usually captured with cameras or mobile phones aligned with the direction of gravity.
Moreover, smart devices are equipped with IMU sensors that can measure the gravity vector accurately. 	
Using the gravity direction, we can compute the roll and pitch angles and use them to align the $y$-axes of the cameras. Let us denote the known rotation matrices used for the alignment of the two cameras as $\M R_1$ and $\M R_2$. In this case,~\eqref{eq:basic_relation} can be written as
\begin{equation}
d_{2i} \M R_2 \M K{_2^{-1}} \M p_{2i}^u = d_{1i} \M R_y \M R_1 \M K{_1^{-1}} \M p_{1i}^u , \label{eq:basic_rot}
\end{equation}
where $\M R_y$ is the unknown rotation from the yaw angle (the unknown rotation around the $y$-axis).

For most of the modern CCD or CMOS cameras, it is reasonable to assume that the camera has square-shaped pixels, and the principal point coincides with the image center~\cite{hartley2012efficient}. In such case, the calibration matrices have the form  
$\M K{_1}={\rm diag}(f_1,f_1,1)$ and $\M K{_2}={\rm diag}(f_2,f_2,1)$.
Relationship~\eqref{eq:basic_rot} between the undistorted image points $\M p_{2i}^u$ and $\M p_{1i}^u$ can be expressed using a $3 \times 3$ homography matrix $\M H$ as 
\begin{equation}
\M p_{2i}^u \sim \M H \M p_{1i}^u , \label{eq:upto_scale}
\end{equation}
where 
$
\M H = \M K_2 \M R_2^{\top} \M R_y \M R_1 \M K{_1^{-1}}
$,
and $\sim$ indicates the equality up to a scale factor. The scale, which is equal to $d_{1i} \over d_{2i}$ can be eliminated by multiplying~\eqref{eq:upto_scale} with the skew symmetric matrix $[\M p_{2i}^u]_\times$ resulting in 
\begin{equation}
[\M p_{2i}^u]_\times \M H \M p_{1i}^u =0 . \label{eq:skew}
\end{equation}
Moreover, equation~\eqref{eq:skew} can be multiplied by $f_1 \over f_2$ resulting in 
\begin{equation}
[\M p_{2i}^u]_\times \M G \M p_{1i}^u =0 , \label{eq:skewG}
\end{equation}
where
\begin{equation}
\M G = \widetilde{\M K}_2 \M R_2^{\top} \M R_y \M R_1 \widetilde{\M K}{_1^{-1}} , \label{eq:G}
\end{equation}
with $\widetilde{\M K}_2 = {1 \over f_2}\M K_2 = {\rm diag}(1,1,w_2)$, $w_2 = {1 \over f_2}$ and $\widetilde{\M K}{_1^{-1}} = f_1 {\M K}_1^{-1} = diag(1,1,f_1)$. 
Since $\widetilde{\M K}_2$ contains unknowns only in its last row and column,
one equation from~\eqref{eq:skewG} does not contain $w_2 = {1 \over f_2}$ and $\lambda_2$.

Rotation matrix $\M R_y$ can be parameterized using the Cayley parameterization, which results in a degree-2 polynomial matrix with only one parameter as follows:
\begin{eqnarray}
\M R_y=\frac{1}{1+s^2}\begin{bmatrix}
1-s^2 & 0 & 2s\\
0 & 1+s^2 & 0\\
-2s & 0 & 1-s^2
\end{bmatrix}.\label{eq:Ry}
\end{eqnarray}
In this case, $s$ corresponds to  $\tan \frac{\theta}{2}$, and $\theta$ is the yaw angle. Hence, $\cos \theta = \frac{1-s^2}{1+s^2}$ and $\sin \theta = \frac{2s}{1+s^2}$. Since~\eqref{eq:skew} is homogeneous, the scale factor $\frac{1}{1+s^2}$ in~\eqref{eq:Ry} can be ignored. Note that this formulation introduces a degeneracy for a $180^\circ$ rotation. An alternative formulation is to use the cosines and sines formulation, which needs two parameters and produces an extra trigonometric identity constraint. In our experiments, we found that the cosines representation leads to a larger template size for the G-J (Gauss-Jordan) elimination and, also, more solutions. Therefore, we only focus on the Cayley parameterization in this paper. 
Our goal is to estimate the rotation, focal lengths and, potentially, radial distortion parameters of two cameras using the minimal number of point correspondences.

\section{Minimal Solutions}\label{sec:solvers}
\begin{table*}[htbp]
	\newcommand{\tabincell}[2]{\begin{tabular}{@{}#1@{}}#2\end{tabular}}
	\caption{The properties of the proposed gravity-based (gray) and state-of-the-art solvers.}
	\vspace{-0.2in}
	\label{table:1}
	\begin{center}
		\begin{tabular}{lccccccc>{\columncolor{mygray}}c>{\columncolor{mygray}}c >{\columncolor{mygray}}c >{\columncolor{mygray}}c}
			\toprule
			 & H4 & H5$\lambda$ & H2$f$ &  H3$f_1 f_2$  & H3$\lambda$  & H5$\lambda_1 \lambda_2$ & H6$\lambda_1 \lambda_2$ & H1$f$ & H2$f_1f_2$ & H2$\lambda$  & H3$\lambda_1 \lambda_2$\\
			\midrule
			Reference & \cite{hartley2003multiple} & \cite{fitzgibbon2001simultaneous}  & \cite{brown2007minimal} & \cite{brown2007minimal} & \cite{jin2008three}\cite{ byrod2009minimal}  & \cite{kukelova2015radial} & \cite{kukelova2015radial} &    &  & & \\
			Different $f$ &\checkmark & \checkmark  &  & \checkmark &   & \checkmark & \checkmark &    & \checkmark &   & \checkmark \\
			Radial distortion & & \checkmark  &   &  & \checkmark & \checkmark & \checkmark & & & \checkmark & \checkmark \\
			Different $\lambda$ &  &   &  & & & \checkmark & \checkmark & & & & \checkmark \\
			No. of points & 4 & 5 &  2 & 3 & 3  & 5 & 6 & 1 & 2 & 2 & 3 \\
			No. of solutions &1 & 18 & 3 & 7 & 18  & 5 & 2 & 4(2) & 4(2) & 6(2) & 6(2) \\ 
			Gravity prior &  &   &  & & &  & & \checkmark &\checkmark &\checkmark &\checkmark \\
			\bottomrule
		\end{tabular}
	\end{center}
	\vspace{-0.2in}
\end{table*}

In this section, we present new minimal solutions, all considering that the images are aligned by the gravity direction, to four practical camera configurations. 

\subsection{Common Focal Length Solver - ${\bf H1f}$}
A simple case arises when the two cameras have the same focal length, \ie, $ f_1 = f_2 = f$, and the cameras have known or zero radial distortion, \ie, $\lambda_1 = \lambda_2 = 0$.  
This happens when the images are captured by a fixed camera with negligible distortion, \ie, narrow field-of-view. 
Also, it can be used for images captured by some wide-angle smartphone cameras which often are undistorted automatically.

This is a 2-DOF problem with respect to $\{s,f\}$. Since each correspondence gives two linearly independent constraints, we only need a single one to solve this problem. Given one point correspondence, constraint~\eqref{eq:skewG} leads to three equations from which only two are linearly independent.  
These equations have the following form
\footnotesize 
\begin{eqnarray}
\bm{a_1}\cdot[s^2fw,s^2f,s^2w,s^2,sfw,sf,sw,s,fw,f,w,1]^\top=0,& 
\normalsize
\label{eq:eq1} \\
\bm{a_2}\cdot[s^2fw,s^2f,s^2w,s^2,sfw,sf,sw,s,fw,f,w,1]^\top=0,& \label{eq:eq2} \\
\bm{a_3}\cdot[s^2f,s^2,sf,s,f,1]^\top=0,& \label{eq:eq3}
\end{eqnarray}
\normalsize
where $\bm{a_1}, \bm{a_2}, \bm{a_3}$ are coefficient vectors that can be computed from the point correspondence and the gravity direction and $w = 1 / f$. 
Note, that equations~\eqref{eq:eq1} and~\eqref{eq:eq2} can be multiplied with $f = 1 / w$ to obtain polynomial equations in two unknowns  $\{s,f\}$ and 9 monomials $\{s^2f^2,s^2f,s^2,sf^2,sf,s,f^2,f,1\}$.
Using~\eqref{eq:eq3}, $f$ can be expressed as a rational function in $s$.  Substituting this expression into~\eqref{eq:eq1} or~\eqref{eq:eq2} and multiplying it with the denominator gives us an univariate polynomial in $s$ of degree 6. This polynomial has, however, always the factor $1+s^2$ that can be eliminated. In this case, we can obtain a quartic equation in $s$, which can be solved in closed-form. Finally, there are up to 4 possible solutions.

\subsection{Varying Focal Length Solver - ${\bf H2f_1 f_2}$}
The next case that we consider assumes unknown different focal lengths. This case occurs, \eg, when the images are taken with a zooming camera. It is a 3-DOF problem with respect to unknowns $\{s,f_1,f_2\}$, and we need at least 1.5 point correspondences to solve it. 
In practice, we still need to sample 2 points, but we only need 3 out of  4 linearly independent equations. 
The 4$^{th}$ equation can be used to eliminate geometrically infeasible solutions.
In this case, the constraint~\eqref{eq:skewG} leads to three polynomial equations of the same form as~\eqref{eq:eq1}-\eqref{eq:eq3}, with $w = {1 \over f_2}$ and $f =f_1$. 

Note that equation~\eqref{eq:eq3} is of degree 3 and does not contain the unknown $w$. In this case, two correspondences give us two equations of form~\eqref{eq:eq3} in two unknowns $\{s,f_1\}$.
These equations can be used to solve for $\{s,f_1\}$. 
In this case $f_1$ can be easily eliminated, leading to a quartic equation in $s$, which can be solved in closed-form.
This is done by rewriting two equations~\eqref{eq:eq3} as
\begin{equation}
\M C(s)	\begin{bmatrix}
	f_1  &
	1
	\end{bmatrix}^\top = 0\label{eq:Cf}
\end{equation}
where $\M C (s)$ is a $2\times 2$ coefficient matrix, and the entries of $\M C$ are polynomials of degree 2 in the hidden variable $s$.
Since~\eqref{eq:Cf} has non-trivial solutions, matrix $\M C(s)$ must be rank-deficient.
Thus, ${\rm det}(\M C(s))=0$, which is a univariate polynomial equation in $s$ of degree 4. 
Once $s$ is computed, $f_1$ can be extracted from the null space of matrix $\M C(s)$.
Then substituting solutions for $\{s,f_1\}$ into~\eqref{eq:eq1} yields up to 4 possible solutions to $w = {1 \over f_2}$.

\subsection{Common Focal Length, Distortion - ${\bf H2\lambda }$}

Another practical case occurs when the images are taken by a camera with fixed unknown focal length and radial distortion, \ie $f_1 = f_2 = f$ and $\lambda_1 = \lambda_2 = \lambda$ in \eqref{eq:skewG}. 
In this case, there are three unknowns $\{s,f,\lambda\}$. 
Thus we need 1.5 point correspondence , or, in practice, 2 point correspondences to solve this problem. Actually, from the second point correspondence we will again use only one constraint. 

Two out of the three equations from~\eqref{eq:skewG} are of degree 6 and one is of degree 4, \ie, the equation of degree four corresponds to the last row of matrix $[\M p_{2i}^u]_\times$ that does not contain $\lambda$ and multiplies only the first two rows of $\widetilde{\M K}_2$, which does not contain $f$. 
Therefore, to solve the problem we use equations of degree 6 and 4 from the first correspondence and the equation of degree 4 from the second correspondence.
The system of these three equations in three unknowns $\{s,f,\lambda\}$ can be solved using the \gb 
method. Using the automatic generator 
\cite{larsson2017efficient}, we obtained a solver that performs G-J elimination of a template matrix of size $30\times 38$, and the eigenvalue decomposition of an $8 \times 8$ action matrix, \ie, the system has 8 solutions. In practice, we found that there are always two imaginary solutions. 
So there are up to 6 possible real solutions. 
An alternative and more efficient way how to solve such system is to  eliminate $f,\lambda$ from the original equations using similar hidden variable trick as in  H2$f_1 f_2$ solver. In this way, we obtain a univariate polynomial in $s$ of degree 8, which can be efficiently solved using Sturm sequences~\cite{gellert2012vnr}. More details on how to eliminate $f,\lambda$ are in the supplementary material.

\subsection{Varying Focal Length, Distortion - ${\bf H3\lambda_1\lambda_2}$}
The final case that we consider, is the most general one, when the focal lengths and distortion parameters of the two cameras are different. 
This is the situation, \eg, when the images are taken using a zooming wide-angle camera. For this problem, there are a total of five unknowns $\{s,f_1,f_2,\lambda_1,\lambda_2\}$. The minimal case is 3 point correspondences, and similar to the 2-point algorithms we only need one equation from the constraints implied by the last one. 

In this case, similarly to the ${\bf H2\lambda }$ solver, two out of the three equations from~\eqref{eq:skewG} are of degree 6 and one is of degree 4, however, now in five unknowns. 
Again only two from these equations are linearly independent.
Moreover, the equation of degree four, \ie, the equation corresponding to the last row of the matrix $[\M p_{2i}^u]_\times$, has a simpler form and contains only three unknowns $\{s,f_1,\lambda_1\}$ as follows:
\begingroup\makeatletter\def\f@size{9}\check@mathfonts
\begin{eqnarray}
&\bm{b} [s^2f_1\lambda_1,s^2f_1,s^2,sf_1\lambda_1,sf_1,s,f_1\lambda_1,f_1,1]^\top=0, \label{eq:eq3r}
\end{eqnarray}
\endgroup
where $\bm{b}$ is a coefficient vector. 
To simplify the solver, we use only this simpler equation~\eqref{eq:eq3r} from the third point correspondence of degree four in three unknowns. In this way, we obtain 5 polynomial equations in 5 unknowns. However, three of them contain only three unknowns $\{s,f_1,\lambda_1\}$. 

Therefore, we can first use the three equations of form~\eqref{eq:eq3r} to solve for $\{s,f_1,\lambda_1\}$. We solve these equations using the hidden variable technique~\cite{cox2006using}. The basic idea of this technique is to consider one of variables as a hidden variable and then compute the resultant. 
We treat $s$ as the hidden variable, \ie, we hide it into the coefficient matrix. 
The equations from the three correspondences can be expressed in terms of the monomials $\{f_1,\lambda_1,f_1,1\}$ as
\begin{equation}
\M C(s)[f_1\lambda_1\ f_1\ 1]^\top = \mathbf 0 , \label{q18}
\end{equation}
where $\M C$ is of size $3\times 3$ with entries that are polynomials in the hidden variable $s$. In this case, $\M C(s)$ has form
\begin{eqnarray}
\M C(s) ={
	\begin{bmatrix}
	\vspace{0.1cm}
	C_1^{[2]}(s) & C_2^{[2]}(s) & C_3^{[2]}(s)\\
	\vspace{0.1cm}
	C_4^{[2]}(s) & C_5^{[2]}(s) & C_6^{[2]}(s)\\
	C_7^{[2]}(s) & C_8^{[2]}(s) & C_9^{[2]}(s)
	\end{bmatrix}},\label{q19}
\end{eqnarray}
where the upper index $^{[\cdot]}$ denotes the degree of the respective polynomial $C_i(s)$. Since~\eqref{q18} has a non-trivial solution, matrix $\M C(s)$ must be rank-deficient. Thus, ${\rm det}(\M C(s))=0$, which is an univariate sextic equation in $s$ 
that can be solved using Sturm sequences.
Once $s$ is known, $f_1,\lambda_1$ can be extracted from the null space of the matrix $\M C(s)$. Finally, $f_2$ and $\lambda_2$ can be extracted from the remaining two equations.

\subsection{Special case}
If the $y$-axis of the camera is considered to be physically aligned with the gravity direction, \ie, $\M R_1, \M R_2$ are identity matrices, 
there are many zero coefficients in systems that appear in our solvers. In such a scenario, all four solvers are very simple and result in one quadratic equation.
The properties of all the stitching solvers are shown in Table~\ref{table:1}.

\section{Experiments}

In this section, we study the performance of the proposed H1$f$, H2$f_1f_2$, H2$\lambda$, H3$\lambda_1 \lambda_2$ solvers on both synthetic and real images. For comparison, the state-of-the-art minimal H2$f$, H3$f_1f_2$ solvers for zero distortion~\cite{brown2007minimal}, the H5$\lambda$~\cite{fitzgibbon2001simultaneous} and H3$\lambda$ solver~\cite{jin2008three,byrod2009minimal} for equal and unknown distortion, and the H5$\lambda_1\lambda_2$ and H6$\lambda_1\lambda_2$ solvers~\cite{kukelova2015radial} for different and unknown distortions are used. 

\subsection{Synthetic Evaluation}\label{general}

\begin{figure*}
	\centering
	\vspace*{-0.15in}
	\hspace*{-0.1in}
	\subfloat[No radial distortion.]{\label{syn:a1}\begin{tabular}[b]{c}  \includegraphics[trim={1mm 0mm 1mm 0mm},clip,width=1.64in]{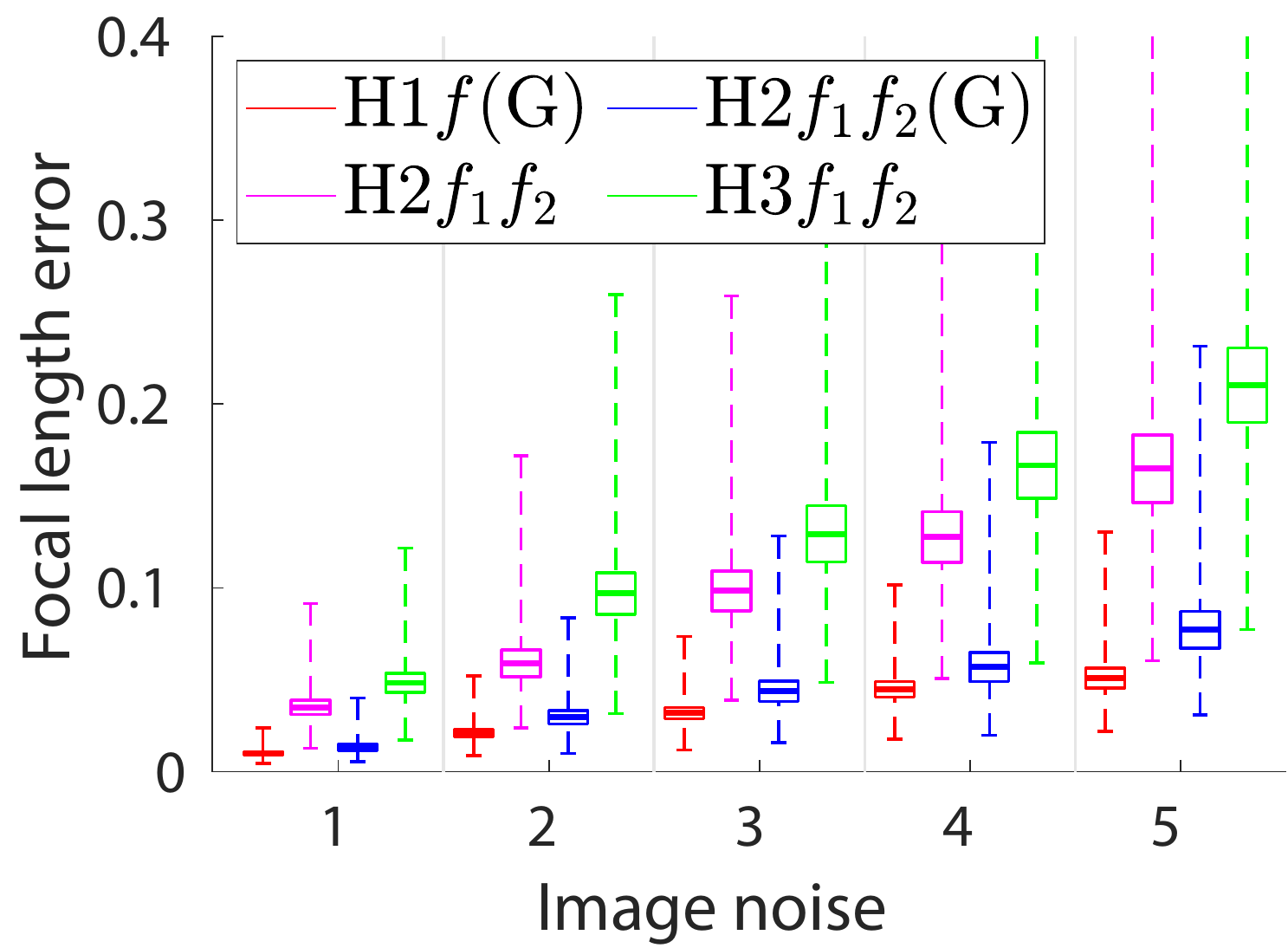}\, \includegraphics[trim={1mm 0mm 1mm 0mm},clip,width=1.64in]{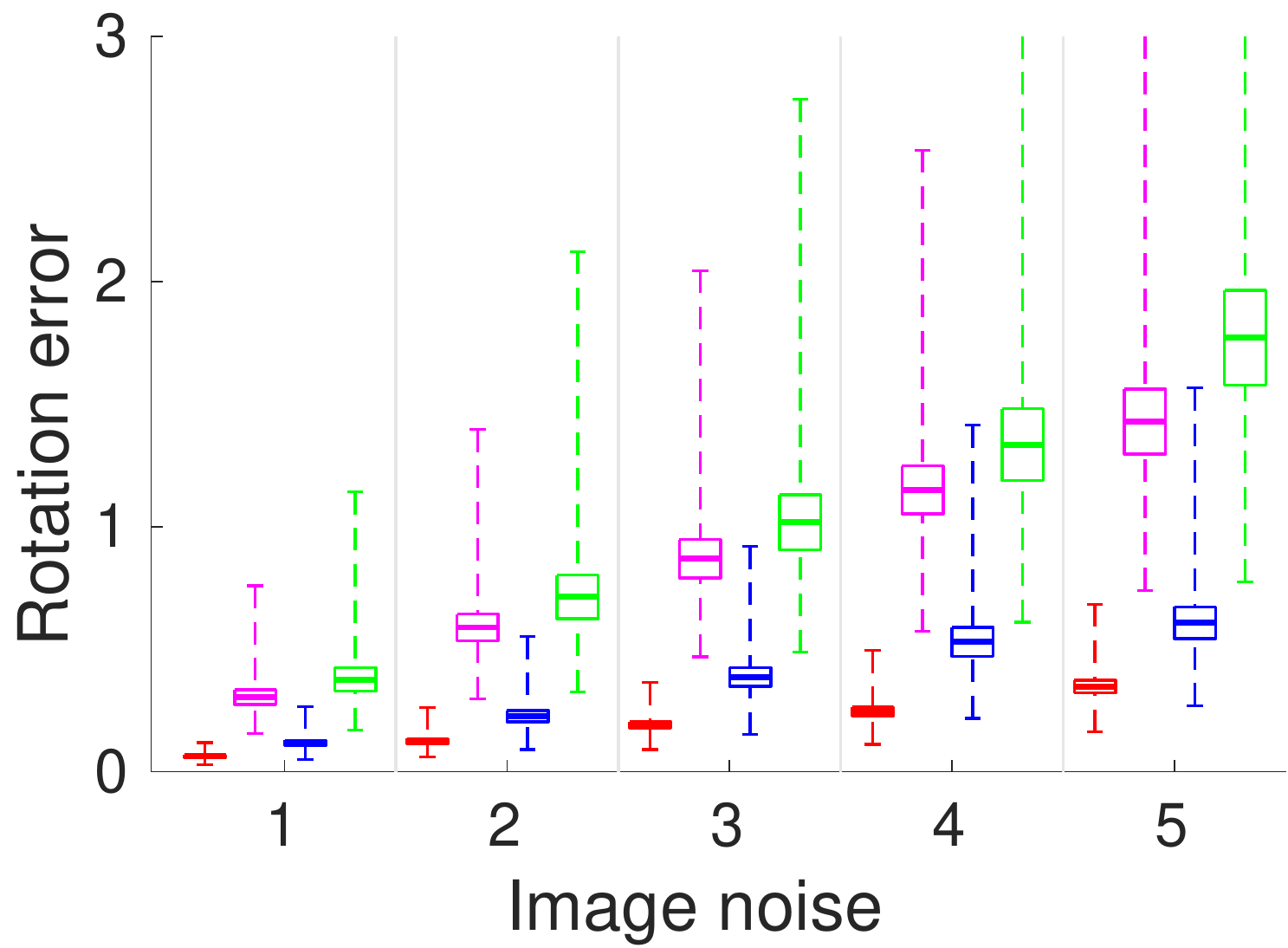}\\
	\includegraphics[trim={1mm 0mm 1mm 0mm},clip,width=1.64in]{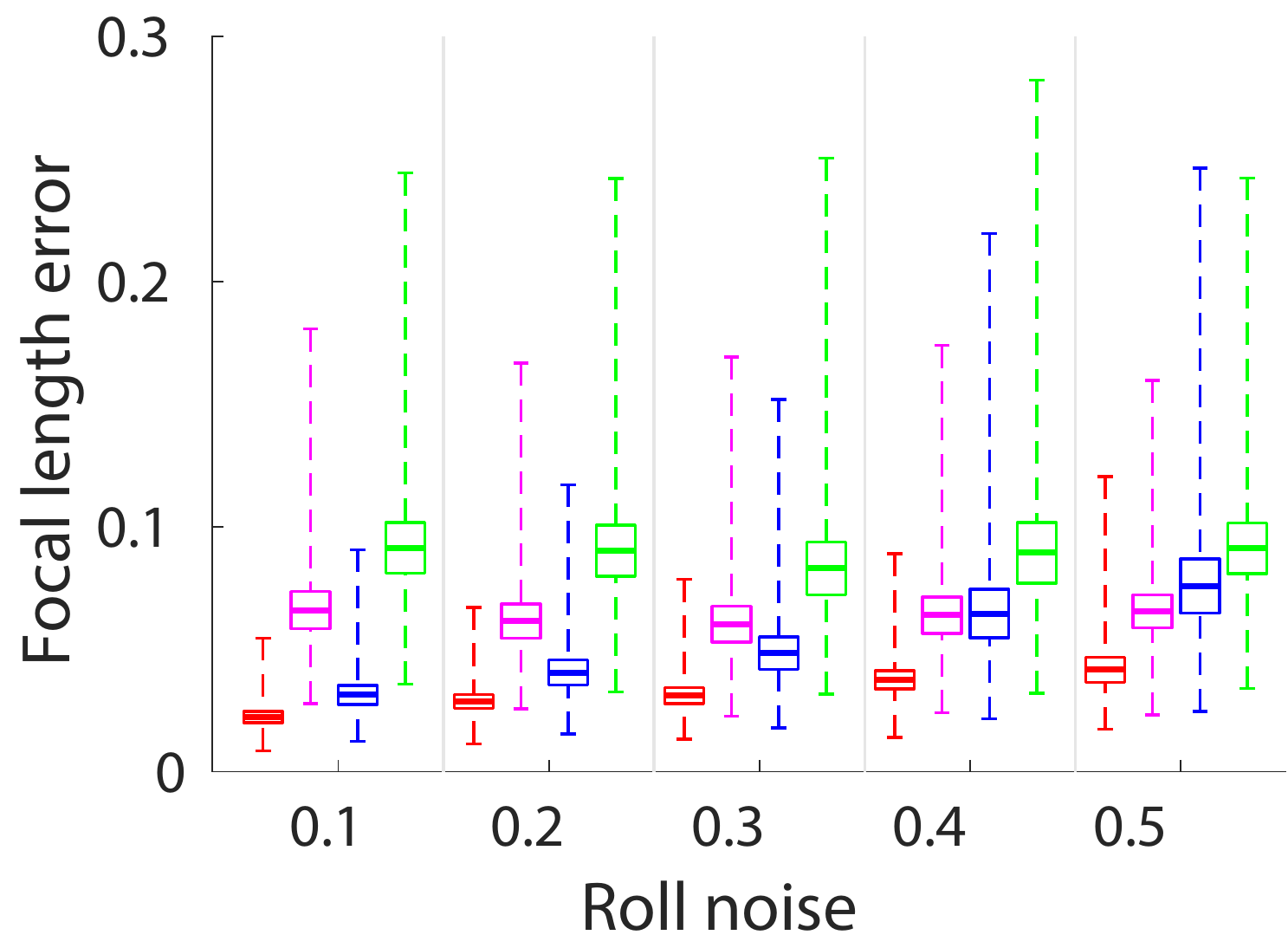}\,
	\includegraphics[trim={1mm 0mm 1mm 0mm},clip,width=1.64in]{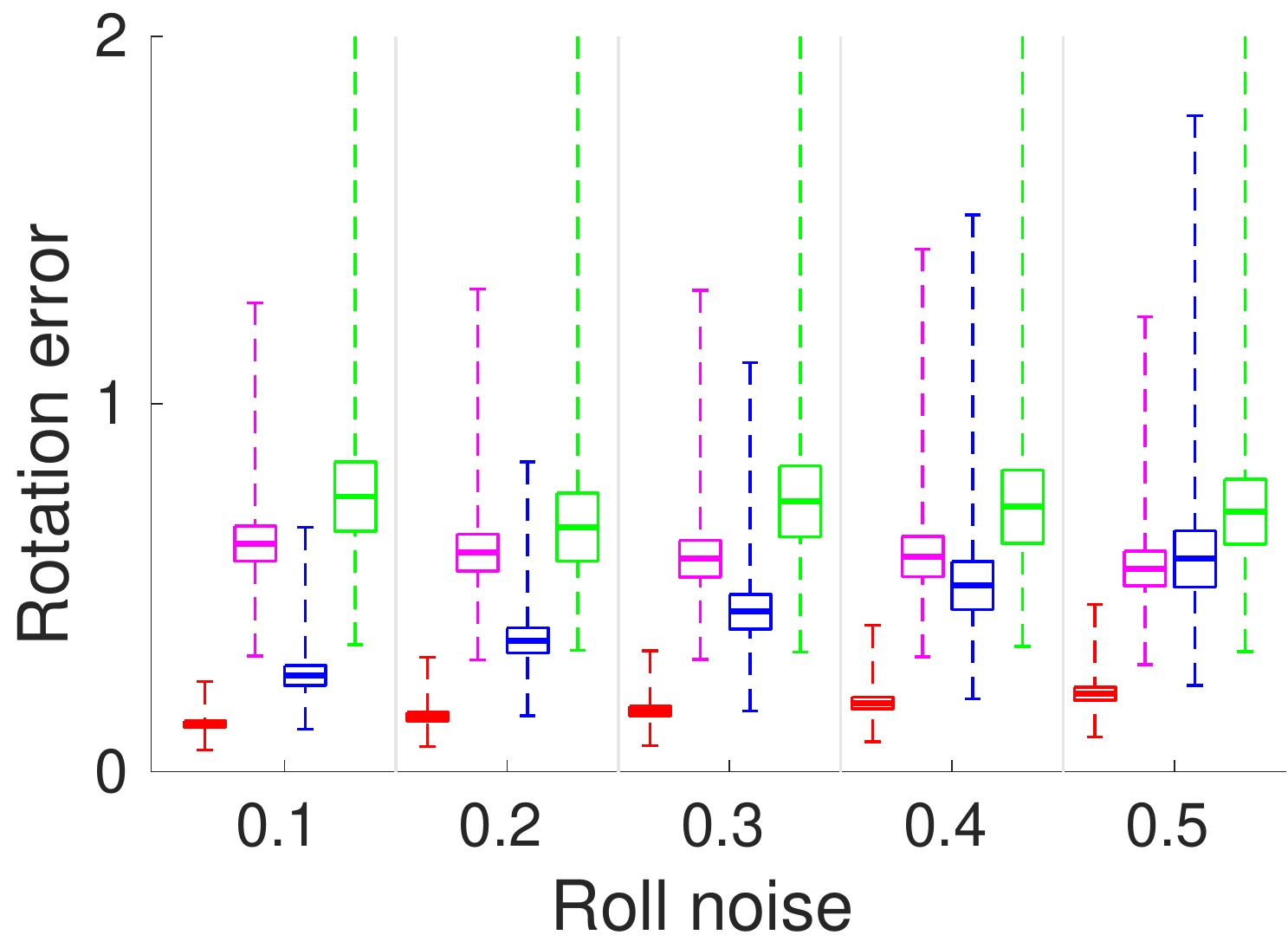}\\
	\includegraphics[trim={1mm 0mm 1mm 0mm},clip,width=1.64in]{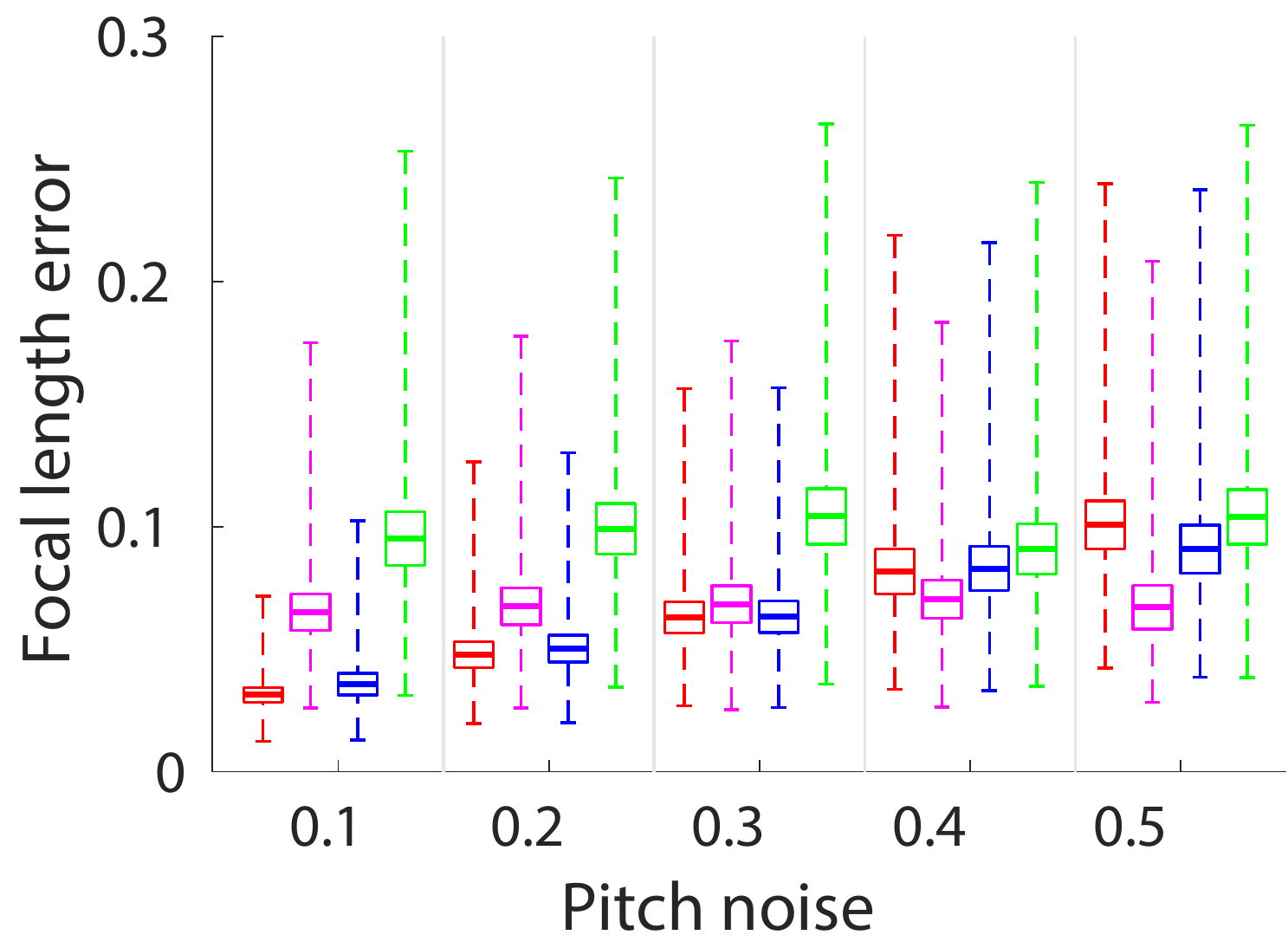}\,
	\includegraphics[trim={1mm 0mm 1mm 0mm},clip,width=1.64in]{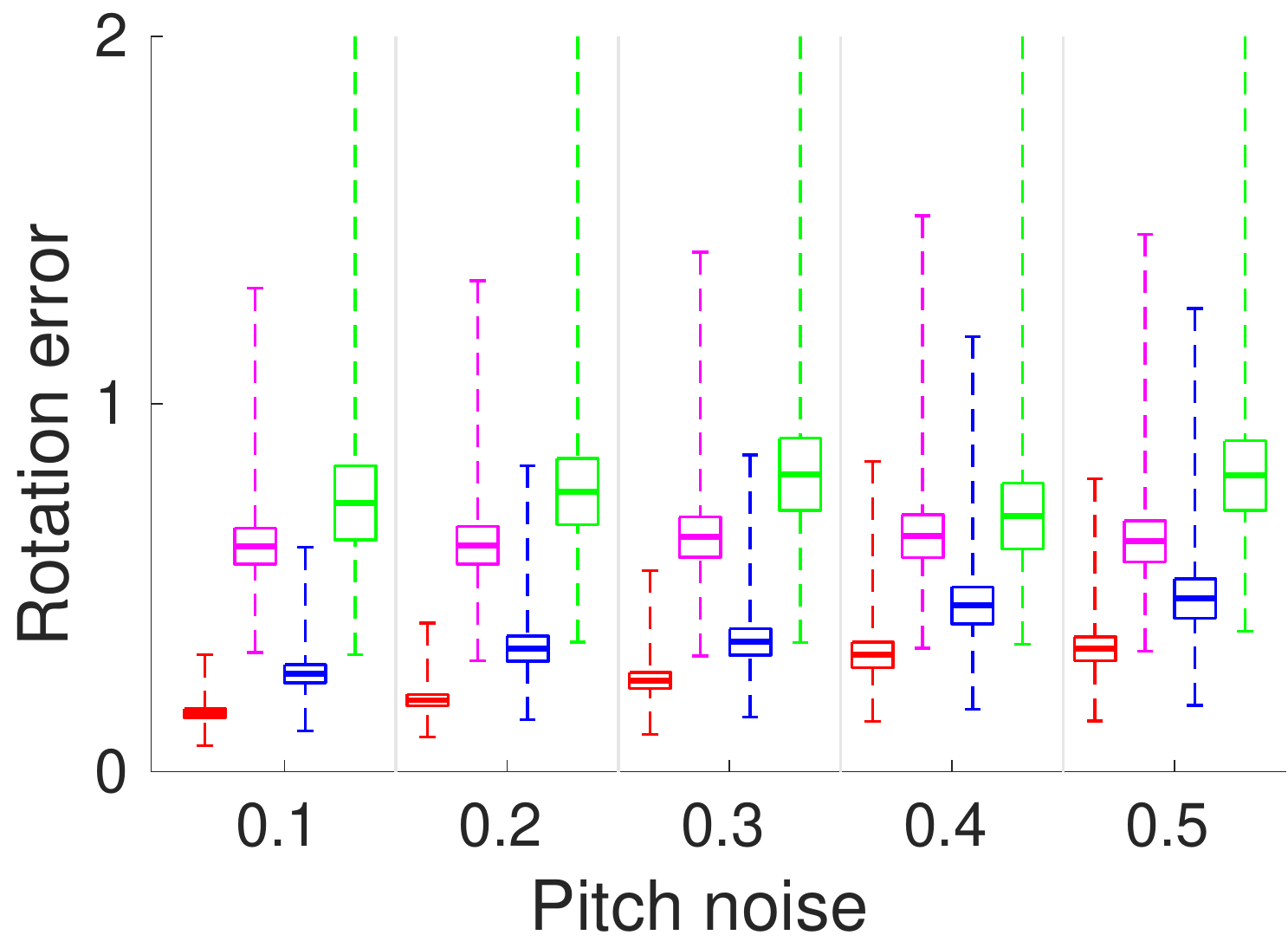}
	\end{tabular}}\hspace*{-0.1in}
	\subfloat[Radial distortion.]{\label{syn:a2}\begin{tabular}[b]{c}  \includegraphics[trim={1mm 0mm 1mm 0mm},clip,width=1.64in]{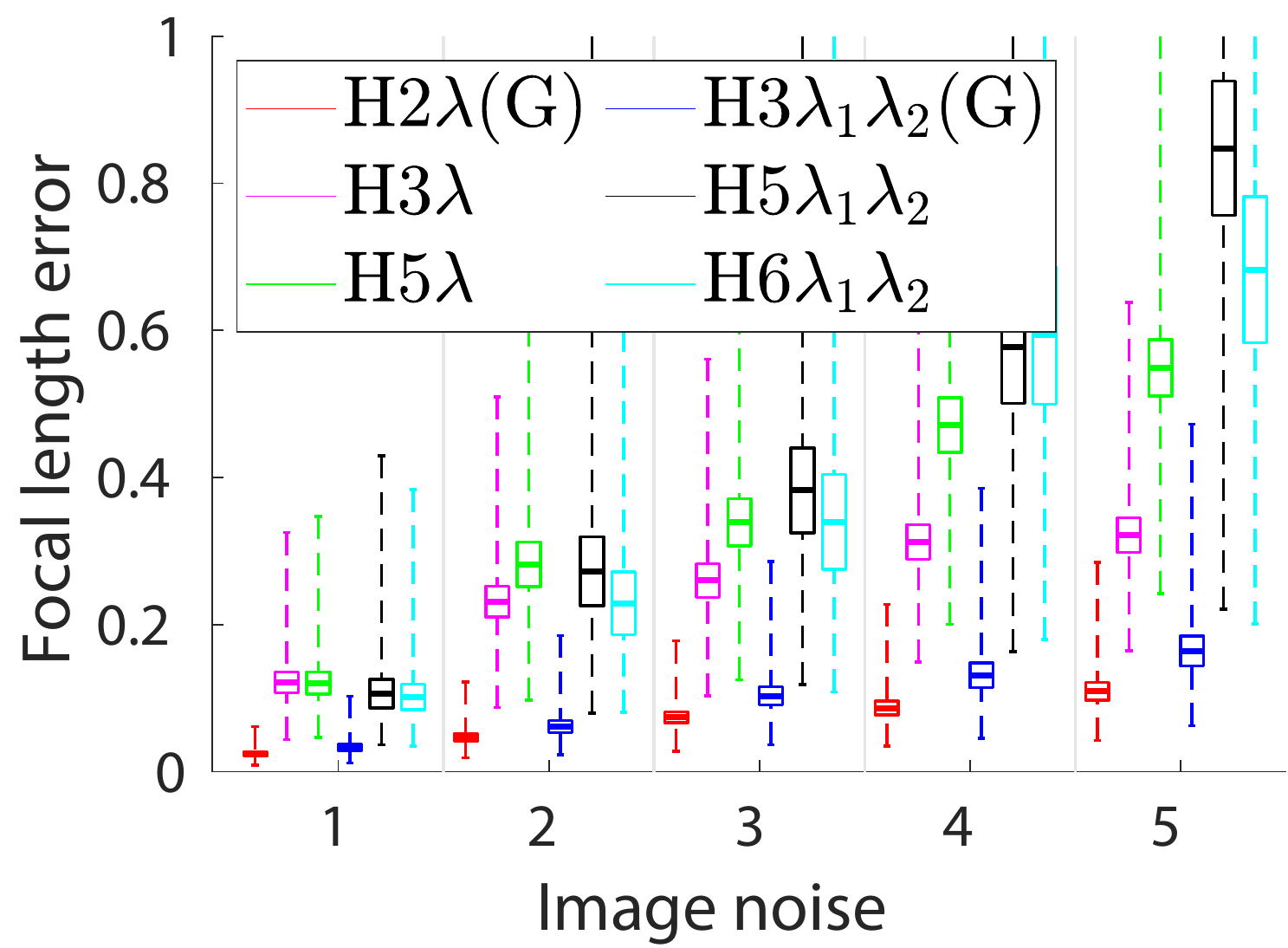}\, \includegraphics[trim={1mm 0mm 1mm 0mm},clip,width=1.64in]{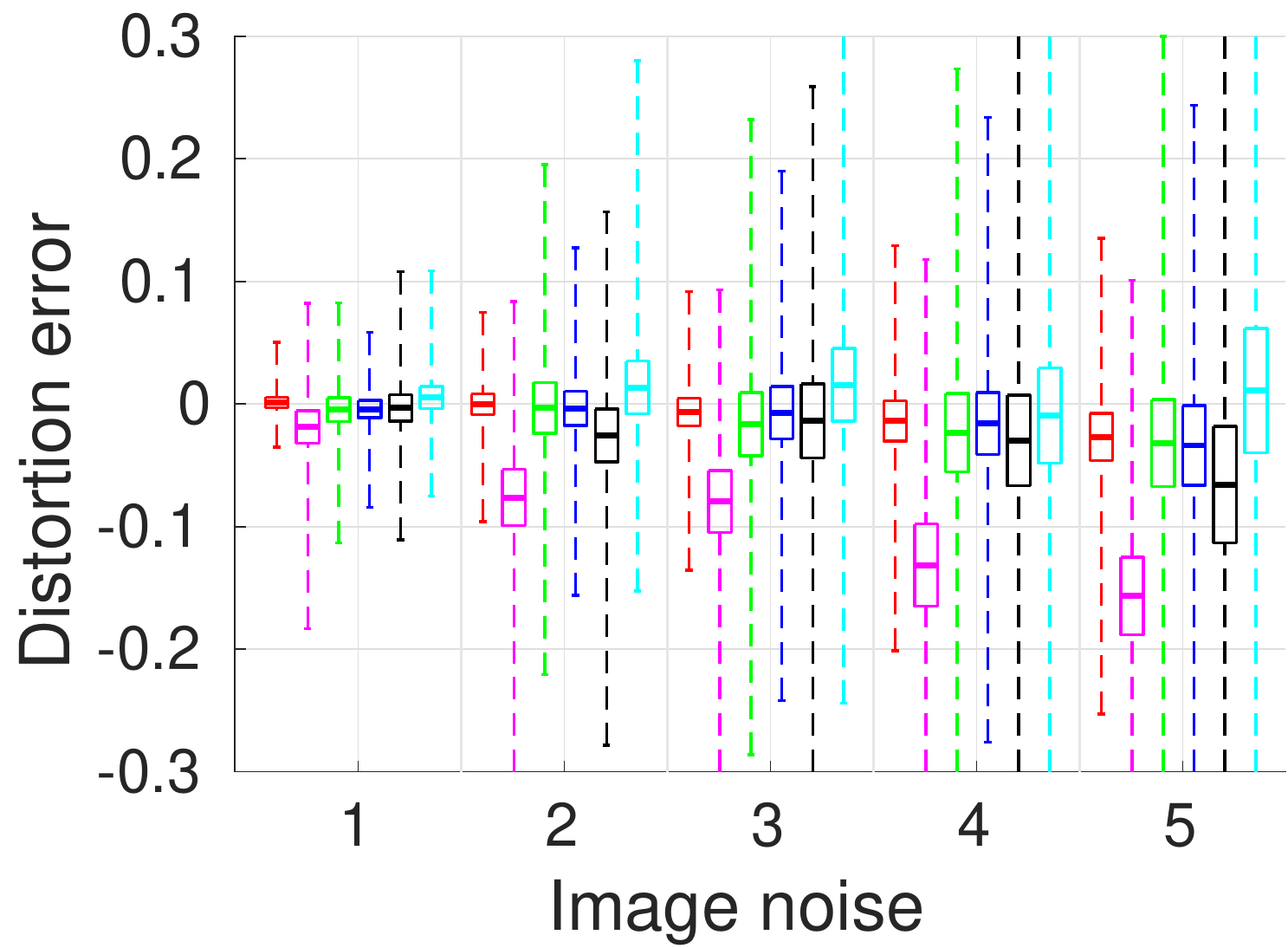}\\
	\includegraphics[trim={1mm 0mm 1mm 0mm},clip,width=1.64in]{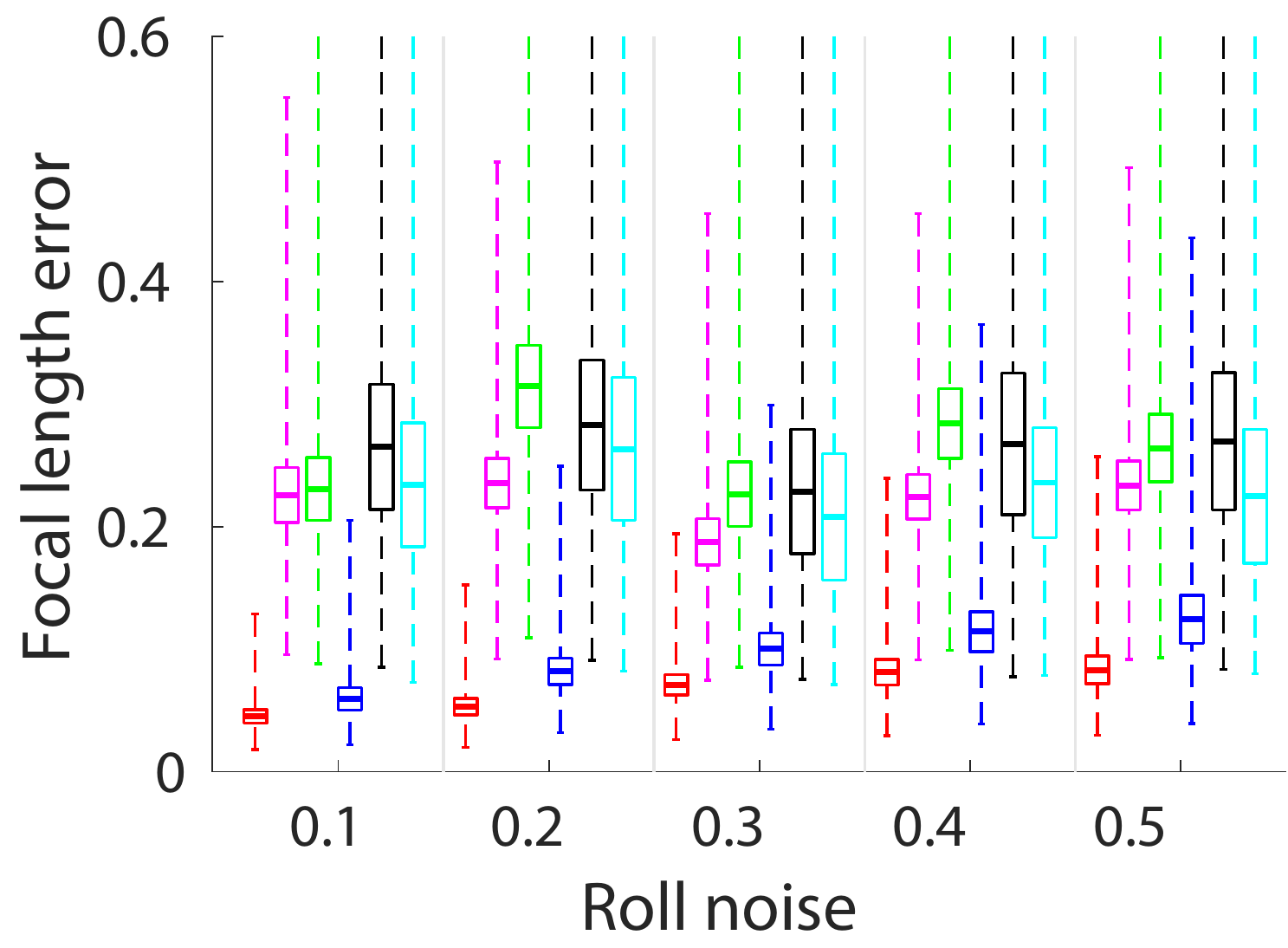}\,
	\includegraphics[trim={1mm 0mm 1mm 0mm},clip,width=1.64in]{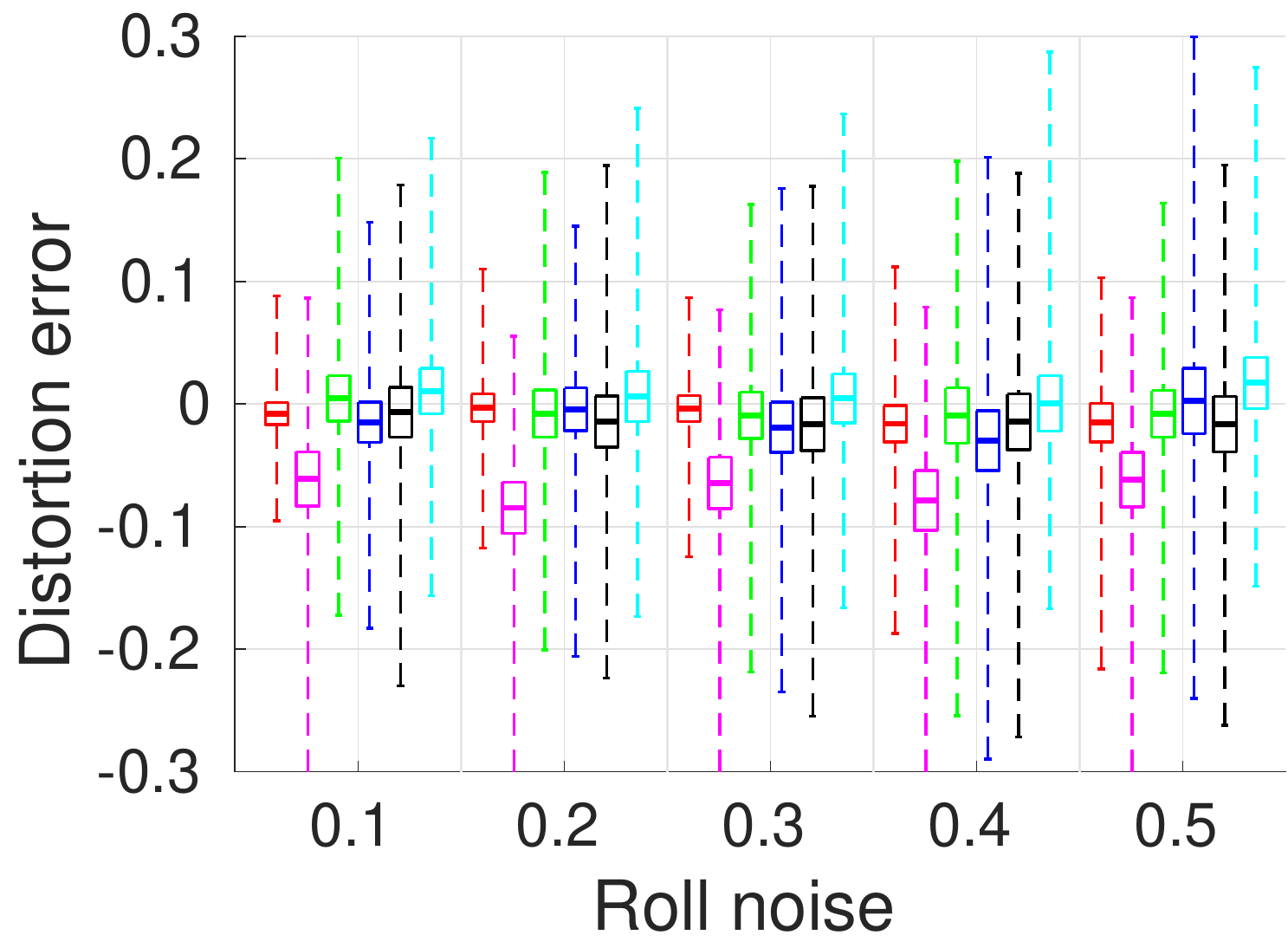}\\
	\includegraphics[trim={1mm 0mm 1mm 0mm},clip,width=1.64in]{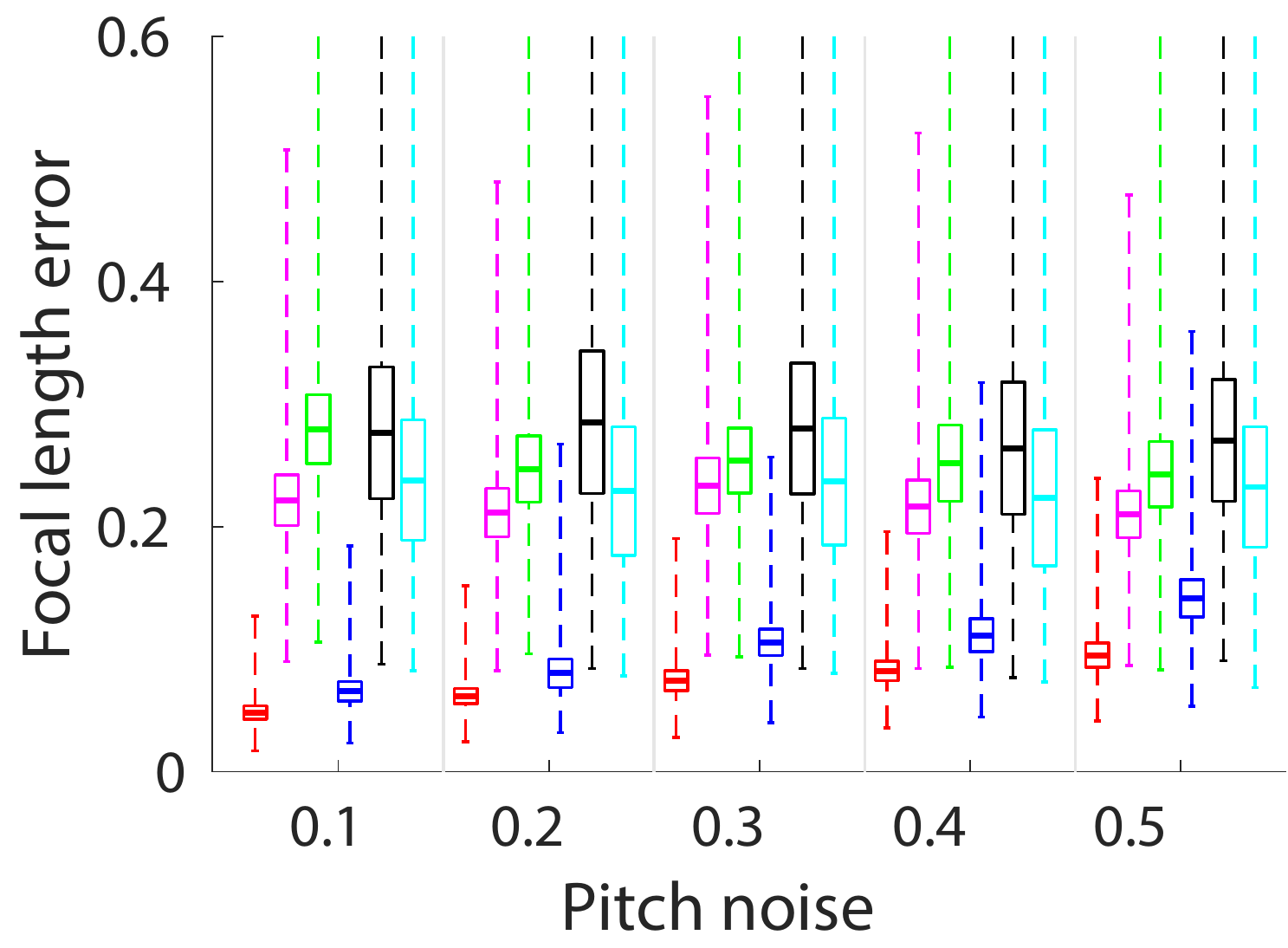}\,
	\includegraphics[trim={1mm 0mm 1mm 0mm},clip,width=1.64in]{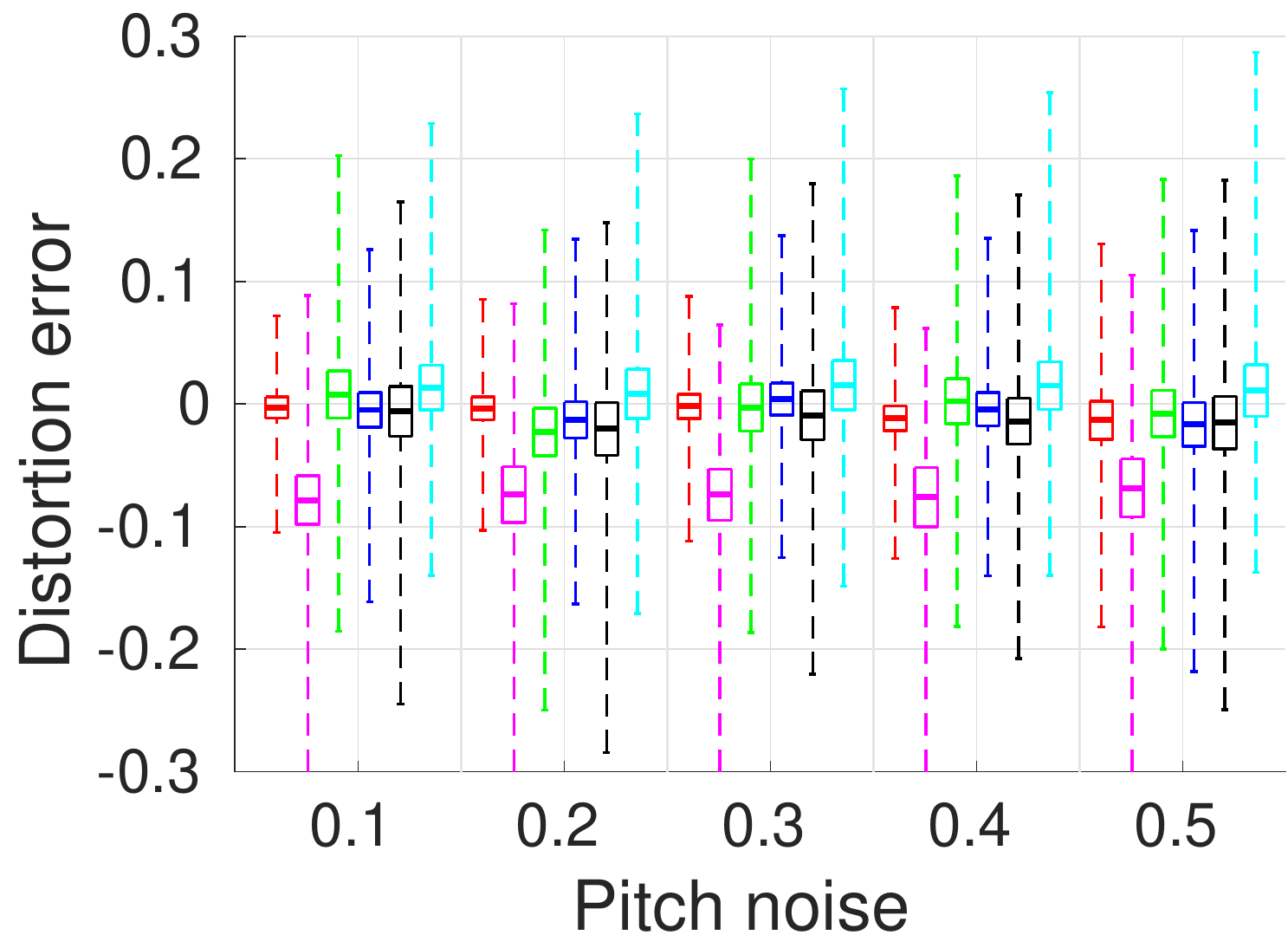}
	\end{tabular}}
	\caption{\textbf{(a)} Boxplot of the focal length error and rotation error for the zero distortion case. \textbf{Left column}: Focal length error. \textbf{Right column}: Rotation error. \textbf{(b)} Boxplot of the focal length error and distortion error for the  distortion solvers. \textbf{Left column}: Focal length error. \textbf{Right column}: Distortion error. From top to bottom: increased image noise, increased roll noise and 2 pixel standard deviation image noise, increased pitch noise with constant 2 pixel standard deviation image noise.}
	\label{syn:a}
\end{figure*}

We choose the following setup to generate synthetic data. 
200 points are randomly distributed in the box $[-3,3]\times[-3,3]\times[4,6]$ in the first camera's local coordinate system. 
A random but feasible rotation was applied to the points. The focal length $f_{g}$ was set to 1,000 pixels. We generated 1,000 different pairs of images with different rotations. The focal length error was defined as $\left|f_e-f_g \right|/f_g$, where $f_e$ is the estimated focal length. For varying focal lengths solvers that return multiple estimates, we compute the focal length error using the geometric mean $f = \sqrt{f_1f_2}$. 

Fig.~\ref{syn:a1} reports the focal length (left column) and rotation error (right column) for the solvers assuming zero distortion. The top row shows the performance under increasing image noise with different standard deviations. The proposed solvers perform significantly better than the other solvers under varying image noise. Since, in real applications, the alignment of the camera or the gravity vector measured by the IMU would not be perfect, we also added noise to the roll and pitch angles to simulate this noisy case. The middle and bottom rows show the performance with increasing roll and pitch noise under constant image noise of 2 pixel standard deviation.
Our solvers are comparable to the state-of-the-art methods even with roll and pitch noise up to 0.5$^\circ$. The upper bound of the noise is chosen to follow the noise from the lower grade MEMS IMUs~\cite{2014Relative}. Nowadays, accelerometers used in modern smartphones and camera-IMU systems have noise levels around $0.06^\circ$ (and expensive “good” accelerometers have $< 0.02^\circ$)~\cite{fraundorfer2010minimal}. 

We also study the performance of the radial distortion solvers. The distortion parameter was set to $\lambda_g=-0.4$, and the error was defined as $\lambda_g-\lambda_e$, where $\lambda_e$ is the estimated distortion. For solvers returning multiple distortions we compute the distortion error using the geometric mean. Fig.~\ref{syn:a2} reports the focal length (left) and radial distortion error (right). 
The H3$\lambda$ solver is very sensitive to noise.

\subsection{Real-world Experiments}

To test the proposed techniques on real-world data, we chose the  \Sun\cite{xiao2012recognizing} panorama dataset\footnote{\url{3dvision.princeton.edu/projects/2012/SUN360/}}. 
The purpose of the \Sun database is to provide academic researchers a comprehensive collection of annotated panoramas covering $360\times180$-degree full view for a large variety of environmental scenes, places and the objects within. To build the core of the dataset, high-resolution panorama images were downloaded and grouped into different place categories. 

To obtain radially distorted image pairs from each 360$^\circ$ panoramic scene, we cut out images simulating a 80$^\circ$ FOV camera with a step size of 10$^\circ$. Thus, the rotation around the vertical axis between two consecutive images is always 10$^\circ$.
Finally, image pairs were formed by pairing all images with a common field-of-view in each scene. 
In total, $579,800$ image pairs were generated. 
Moreover, to test the methods also in cases when there is no distortion, we undistorted the images using the ground truth distortion parameters.
In each image, $8000$ SIFT keypoints are detected in order to have a reasonably dense point cloud and a stable result~\cite{IMC2020}.
We combined mutual nearest neighbor check with standard distance ratio test~\cite{lowe1999object} to establish tentative correspondences~\cite{IMC2020}.
To get ground truth correspondences which can be used to calculate the re-projection error, we composed the ground truth homography and selected its inliers given the used inlier-outlier threshold.

To test the solvers on real-world data, we chose a locally optimized RANSAC, {\ie}, GC-RANSAC~\cite{barath2017graph}. 
In GC-RANSAC (and other locally optimized RANSACs), two different solvers are used: (a) one for estimating the homography from a minimal sample and (b) one for fitting to a larger-than-minimal sample when doing final homography polishing on all inliers or in the local optimization step. 
For (a), the objective is to solve the problem using as few correspondences as possible since the processing time depends \textit{exponentially} on the number of correspondences required for the estimation. 
The proposed solvers are included in this step.
For (b), when testing the methods on the distorted images, we applied the H6$\lambda_1\lambda_2$ solver~\cite{kukelova2015radial} to estimate the homography and distortion parameters from a larger-than-minimal sample.
In case when the undistorted images are used, we applied the normalized H4~\cite{hartley2003multiple} homography estimator.
The inlier threshold was set to $3$ pixels.

The cumulative distribution functions (CDF) of the robust estimation's processing times (in seconds), 
re-projection and focal length errors are shown in Fig.~\ref{fig:real_experiments}. 
Being accurate is interpreted as a curve close to the top-left corner.
The top row reports the results on the distorted and the bottom row shows the results on the undistorted images.

\noindent
\textit{Distorted images.} The run-times of the robust estimation using the proposed H1$f$(G), H2$\lambda$(G) and H2$f_1f_2$(G) solvers are amongst the best ones with H1$f$(G) being the fastest solver.
All these solvers are far real-time, being faster than $0.05$ seconds in $95-99\%$ of the cases.
In terms of re-projection error, solvers not considering the radial distortion have more very accurate results -- their curves go up early. However, the curves of the solvers estimating the radial distortion goes higher -- they are on average more accurate --, as it is expected.
The proposed solvers which estimate the focal length and do not consider radial distortion lead to the most accurate focal lengths. 
The best method, both in terms of processing time and focal length accuracy, is the proposed H1$f$(G) solver.

\noindent
\textit{Undistorted images.} 
After the distortion parameters are used to undistort the images, the most accurate results are clearly obtained by the solvers not considering the radial distortion. 
While, again, the fastest methods are the proposed ones, they are now the most accurate ones as well both in terms of re-projection and focal length errors.

\begin{figure*}[t]
    \centering
    \includegraphics[width=0.99\textwidth]{assets/legend.pdf}
	\subfloat[The captured smartphone dataset.]{
    \includegraphics[trim={0mm 0mm 8mm 0mm},clip, width=0.325\textwidth]{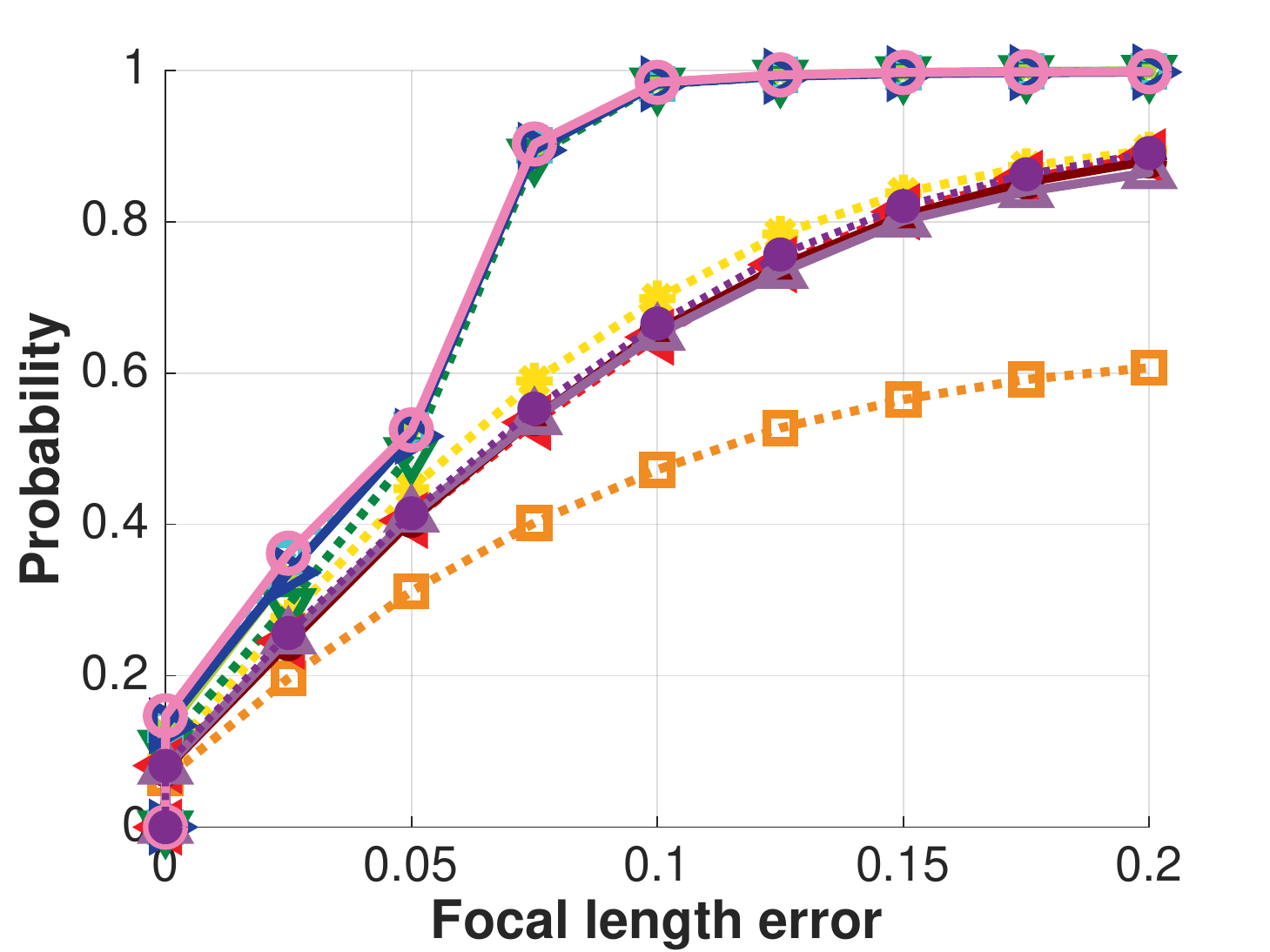}\,
    \includegraphics[trim={0mm 0mm 8mm 0mm},clip, width=0.325\textwidth]{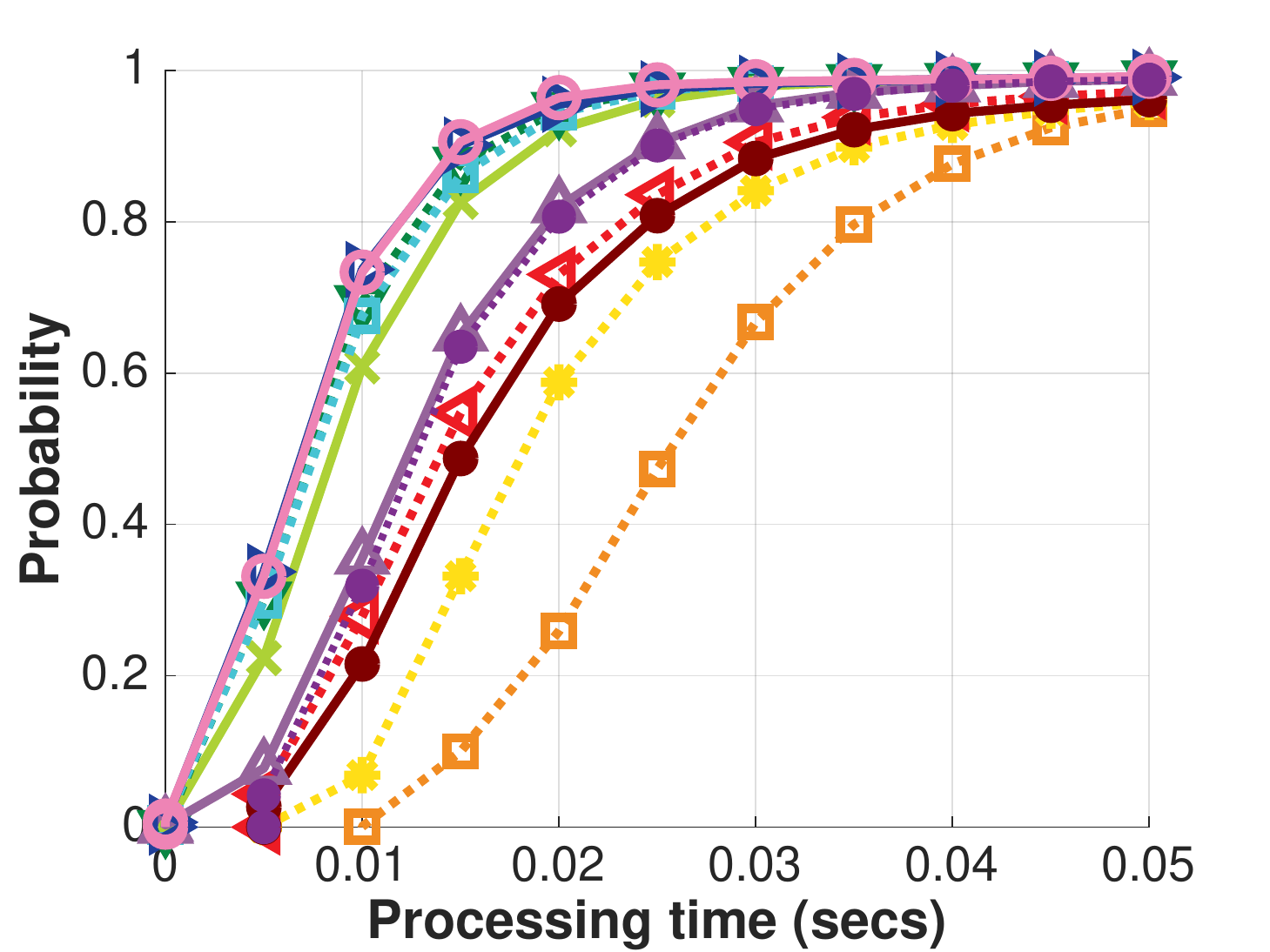}
    \label{fig:phone_experiments}}
	\subfloat[Theoretical RANSAC iterations.]{\includegraphics[trim={0mm 0mm 8mm 0mm},clip,width=0.325\textwidth]{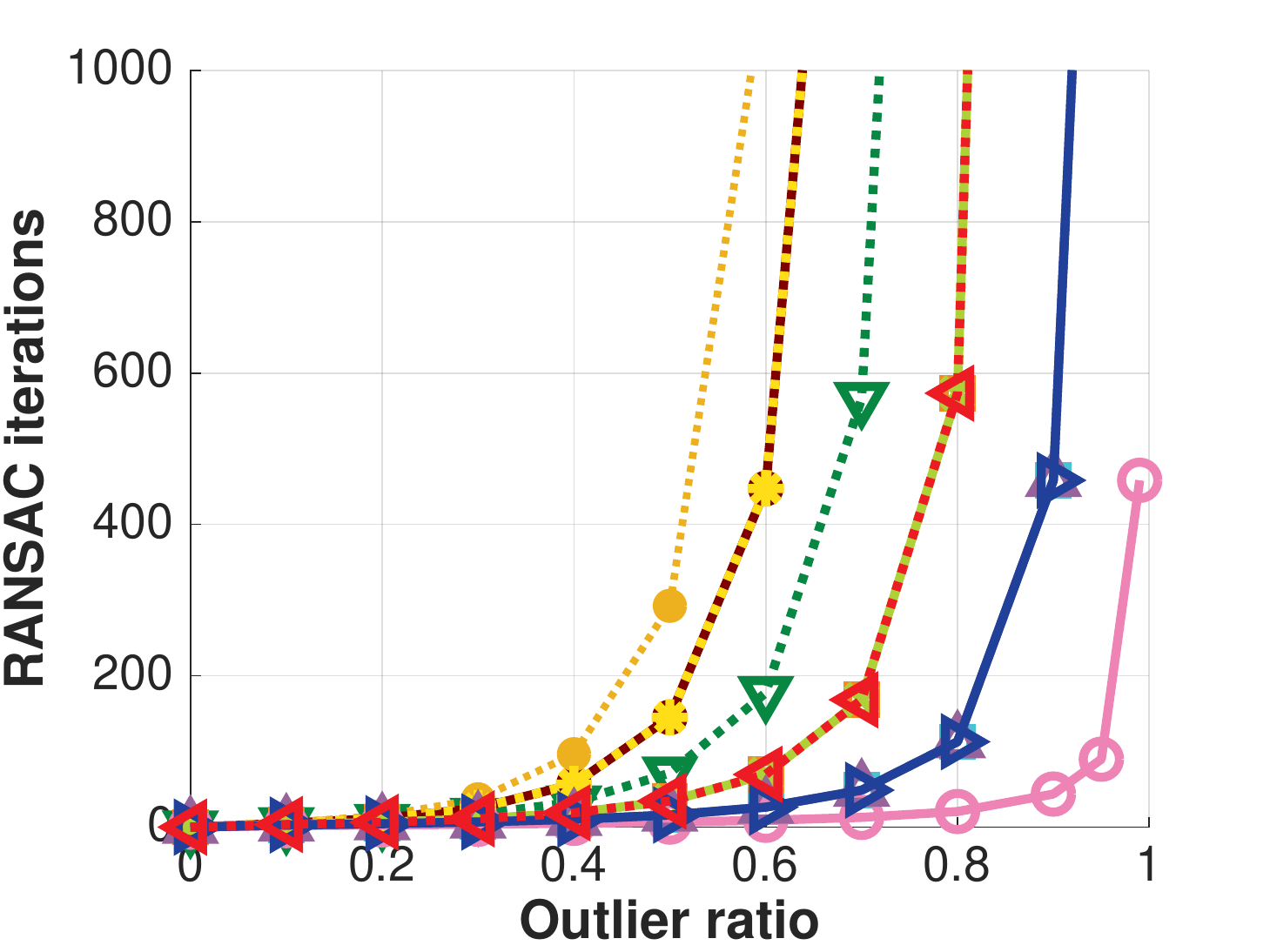}
    \label{fig:ransac_iterations}}
    \caption{ 
        \textbf{(a)} The cumulative distribution functions of the processing time (in seconds) and focal length errors of GC-RANSAC when combined with different minimal solvers. 
        The values are calculated from a total of $7,770$ image pairs from the captured phone dataset. Being accurate is interpreted as a curve close to the top-left corner. 
        \textbf{(b)} The theoretical number of RANSAC iterations for each solver plotted as the function of the outlier ratio. The confidence was set to $0.99$.}
\end{figure*}

\noindent
\textbf{Smartphone images.} 
To further show the benefits of the proposed solvers, we tested them on the data recorded from two devices (iPhone 6s, iPhone 11 pro) with three cameras: the wide-angle cameras of the iPhone 6s and iPhone 11 with focal lengths of 29mm and 26mm, respectively, and the ultra wide-angle camera of the iPhone 11 with a focal length of 13mm. 
The sequences were captured at 1280x720@30Hz with the IMU readings captured at 100Hz. The images and IMU data are synchronized based on their timestamps. $4240$ image pairs for the two wide-angle cameras and $3530$ image pairs for the ultra wide-angle camera with synchronized gravity direction were obtained. 
Since we do not know the ground truth homography and, thus, ground truth inliers, we do not measure the re-projection error on these datasets.
The CDFs of the focal length errors and processing times are shown in Fig.~\ref{fig:phone_experiments}.
The same trend can be seen as before, \ie, the proposed solvers lead to the fastest and most accurate robust estimation.


\begin{figure*}[t]
    \centering
    \includegraphics[width=0.99\textwidth]{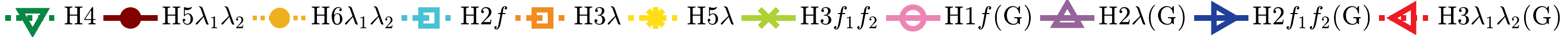}
	\subfloat[Distorted images]{\label{figure:a51} 
	    \includegraphics[trim={1mm 0mm 10mm 0mm},clip, width=0.325\textwidth]{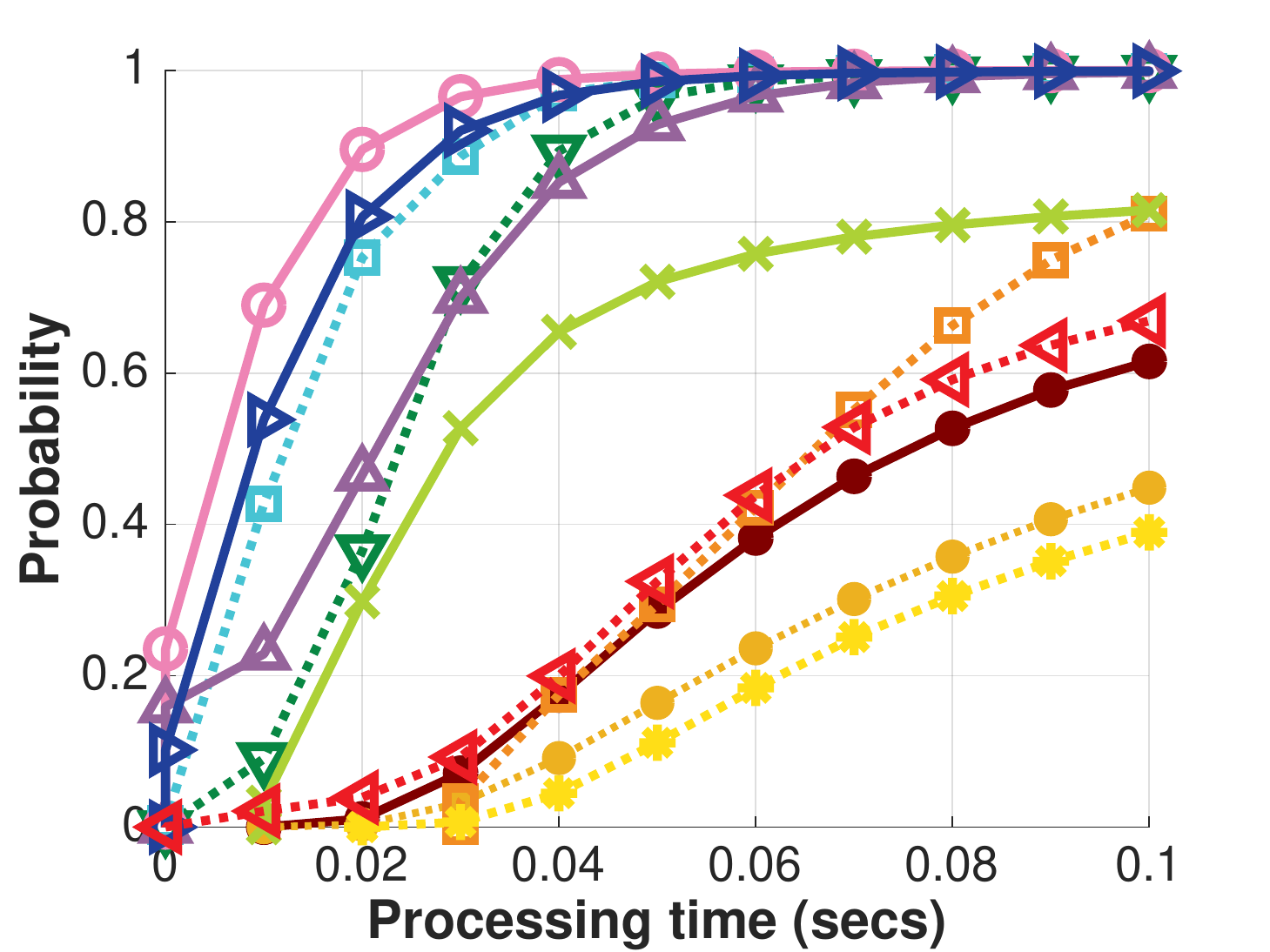}\,
	    \includegraphics[trim={1mm 0mm 10mm 0mm},clip, width=0.325\textwidth]{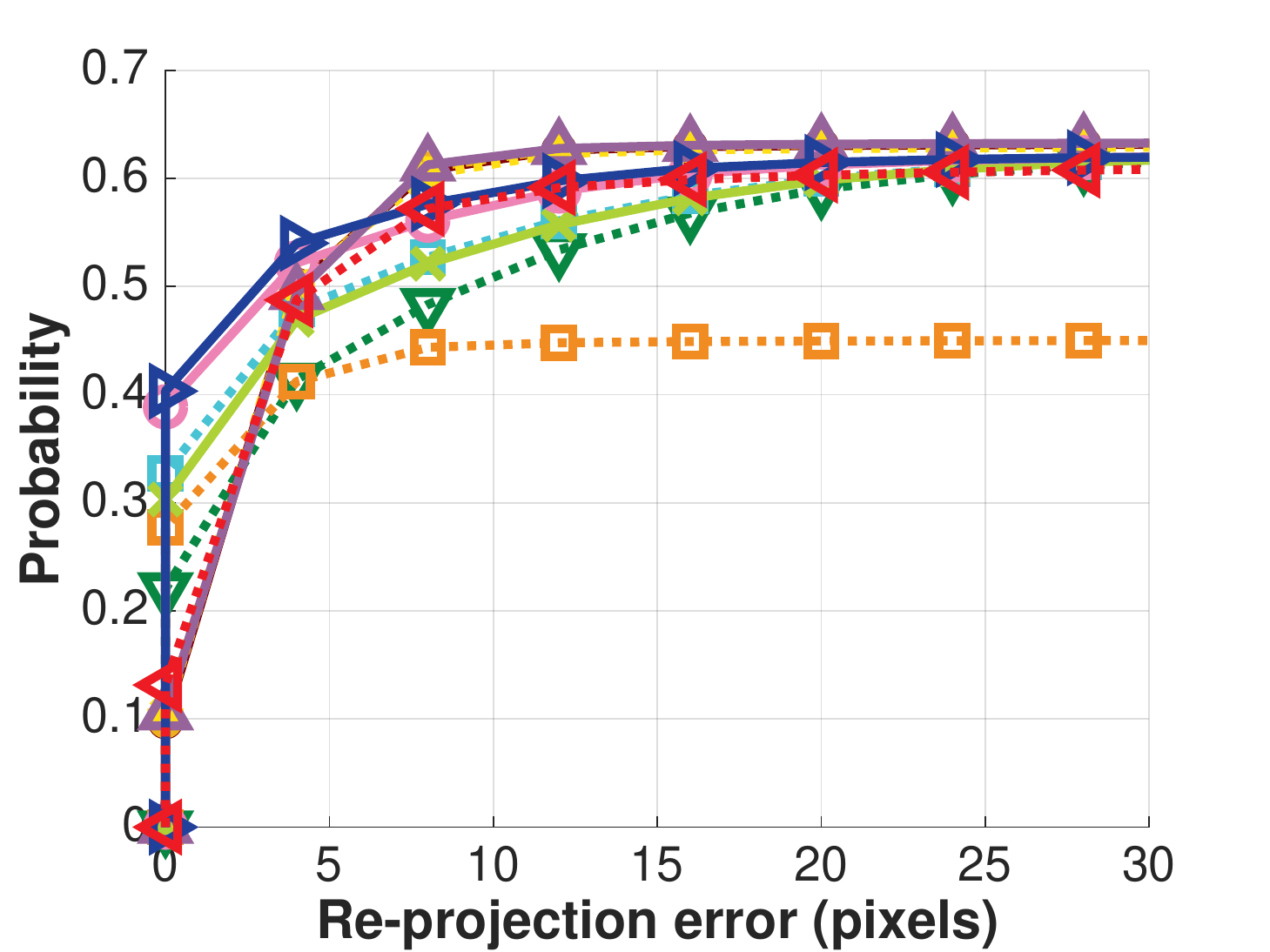}\,
	    \includegraphics[trim={1mm 0mm 10mm 0mm},clip, width=0.325\textwidth]{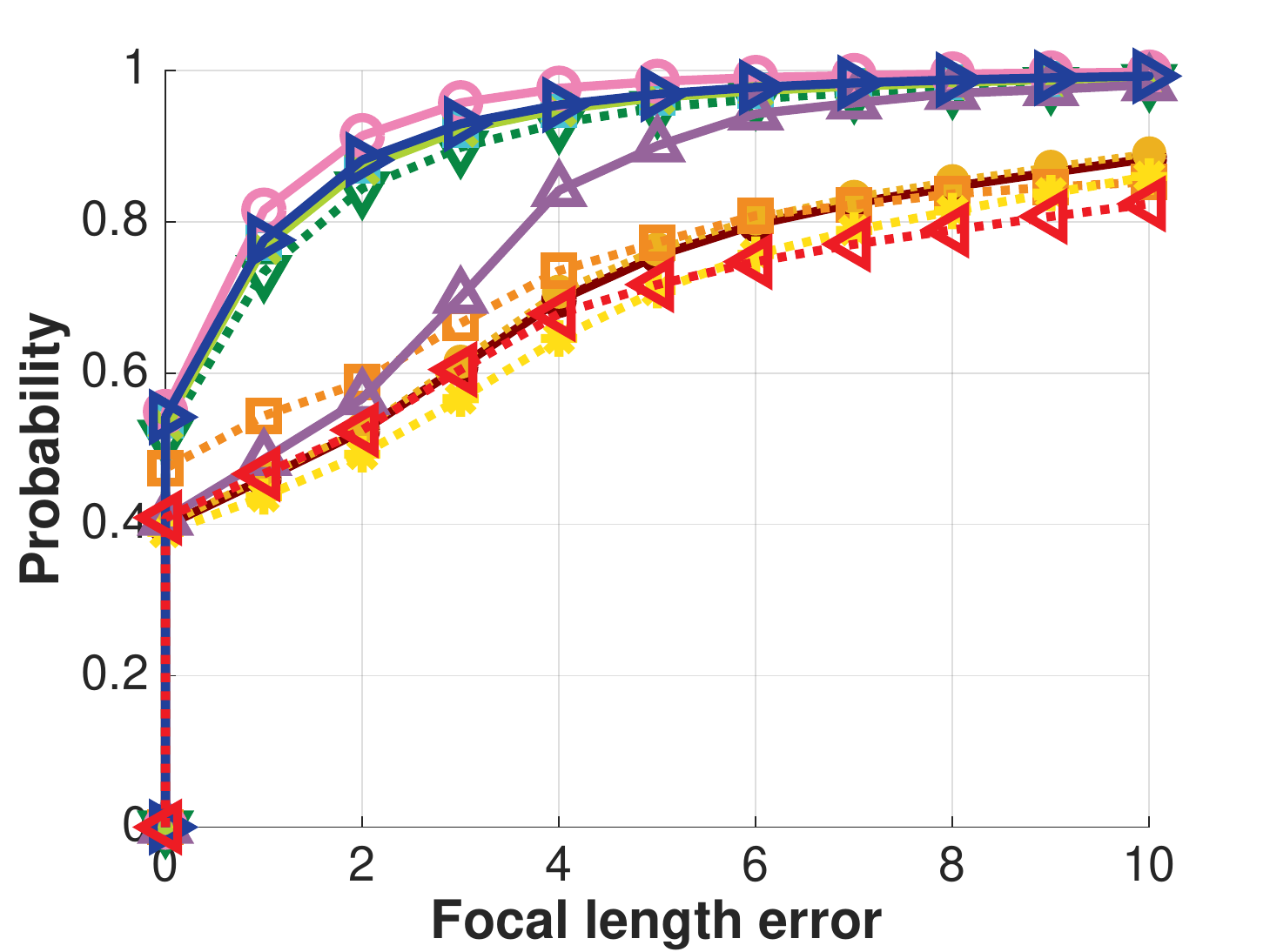}}\\
	\subfloat[Undistorted images]{\label{figure:a51} 
	    \includegraphics[trim={1mm 0mm 10mm 0mm},clip, width=0.325\textwidth]{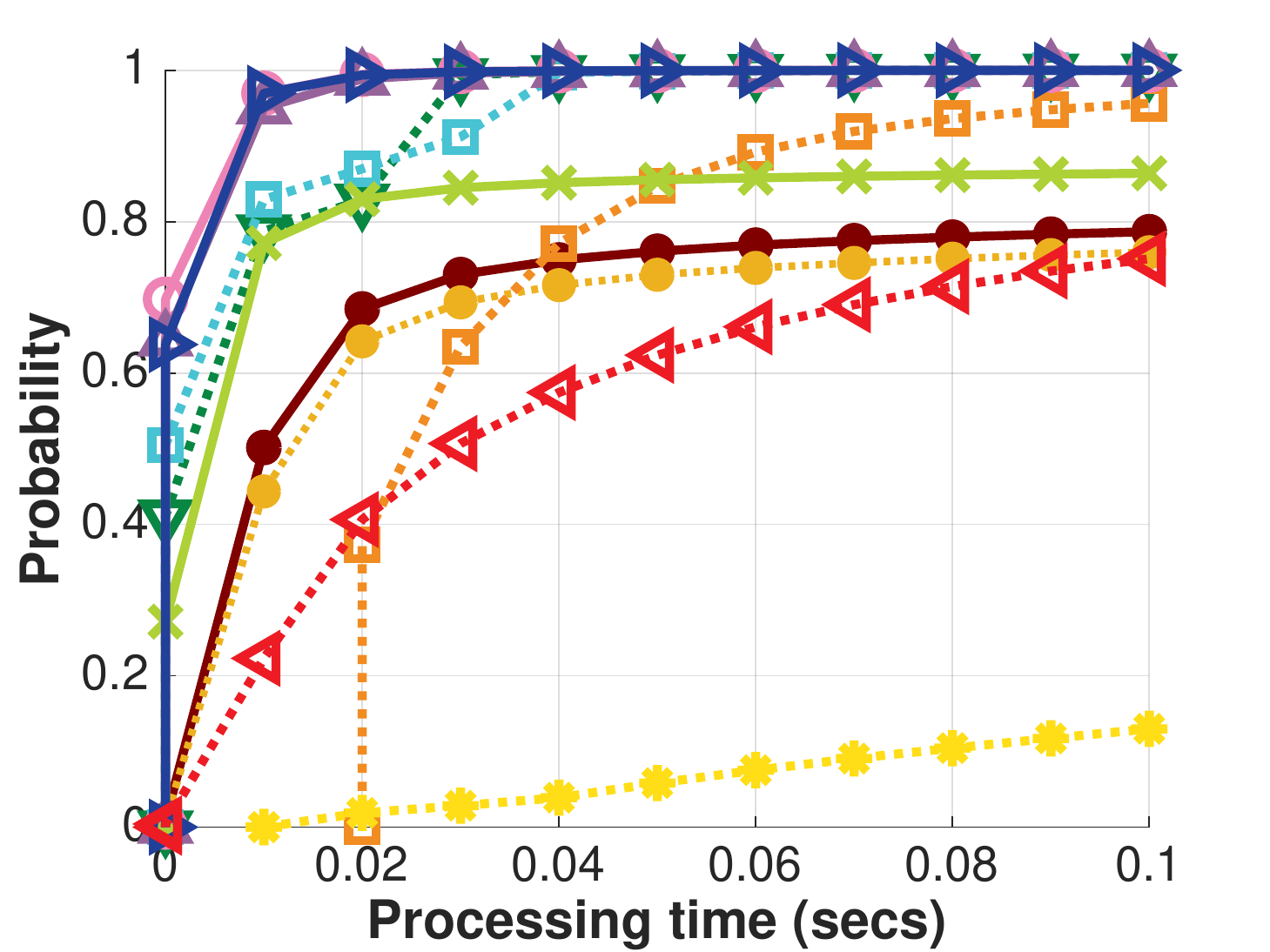}\,
	    \includegraphics[trim={1mm 0mm 10mm 0mm},clip, width=0.325\textwidth]{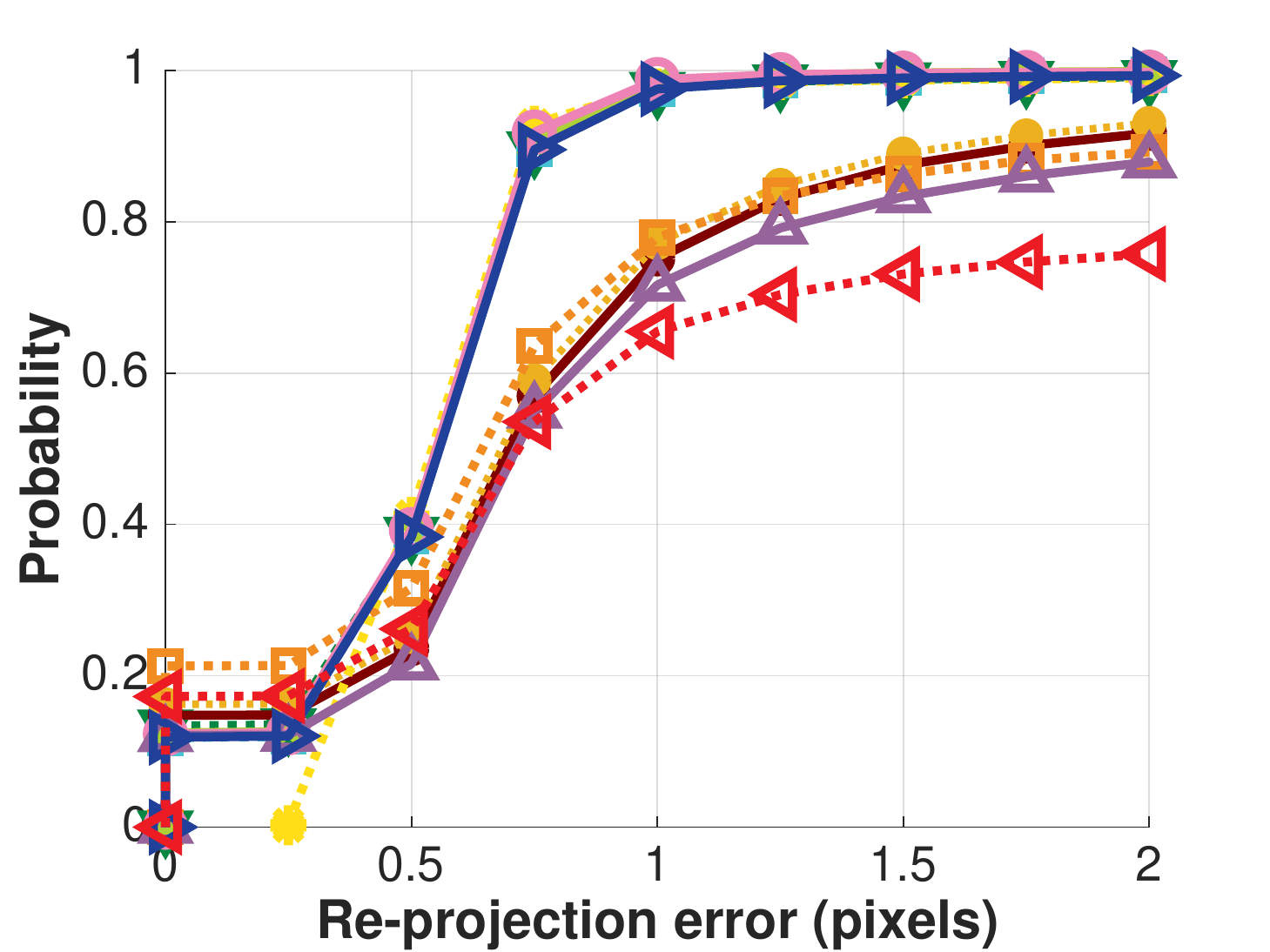}\,
	    \includegraphics[trim={1mm 0mm 10mm 0mm},clip, width=0.325\textwidth]{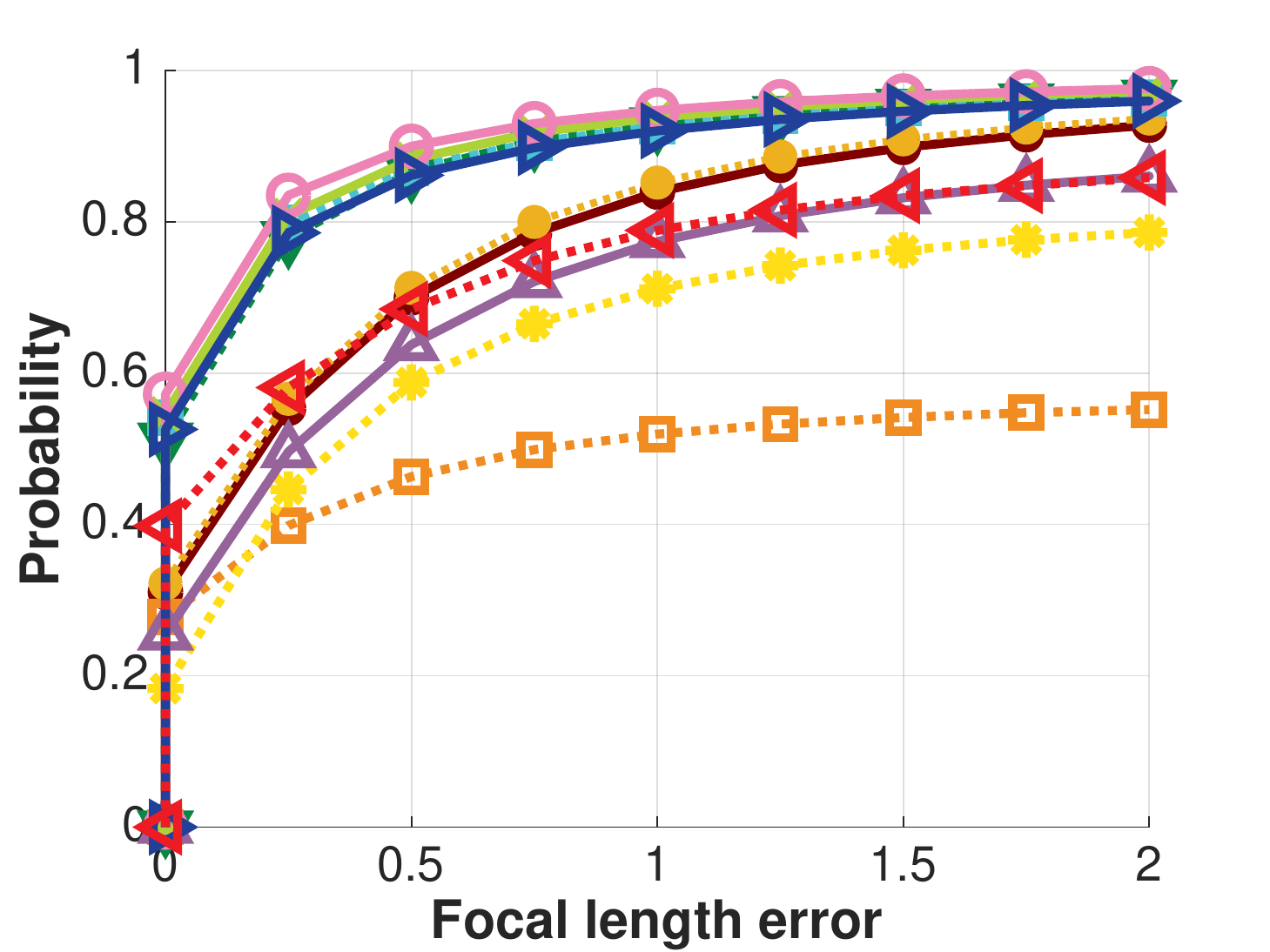}}
    \caption{ 
        The cumulative distribution functions of the processing times (in seconds), average re-projection errors (in pixels) and focal length errors of Graph-Cut RANSAC when combined with different minimal solvers. 
        The values are calculated from a total of $579,800$ image pairs from the \Sun dataset.  
        The confidence was set to $0.99$ and the inlier threshold to $3$ px.
    Being accurate is interpreted as a curve close to the top-left corner. (Top) distorted images. (Bottom) undistorted images.
    }
    \label{fig:real_experiments}
\end{figure*}

\subsection{Computational Complexity}

The computational complexity and run-time of a single estimation of the compared solvers are reported in the following table. We only show the major steps performed by each solver. The number in the cells, \eg $9 \times 18$ denotes the matrix size to which the 
G-J elimination
or Eigen-decomposition is applied. The number in the fourth column denotes the degree of the univariate polynomial which needs to be solved. 
\begin{table}[h]
	\newcommand{\tabincell}[2]{\begin{tabular}{@{}#1@{}}#2\end{tabular}}
	\begin{center}
		\begin{tabular}{lcccc}
			\toprule
			Solver & G-J & Eigen & Poly  &  Time ($\mu$s) \\ \rowcolor{mygray}
			\midrule
			H1$f$(G) & - & - &  4 & 5 \\ \rowcolor{mygray}
			H2$f_1 f_2$(G) & - & - & 4 & 5  \\ \rowcolor{mygray}
			H2$\lambda$(G) & -  & -& 8 & 5   \\ \rowcolor{mygray}
			H3$\lambda_1\lambda_2$(G) &  - & - & 6 & 5 \\
			H2$f$ & - & $3\times3$ & - & 6 \\
			H3$f_1 f_2$ & - & $7\times7$ & - & 7 \\
			H3$\lambda$ & $90\times132$ & $25\times25$ & - & 80 \\
			H5$\lambda$ & $9\times18$ & $18\times18$ & - & 40 \\
			H5$\lambda_1\lambda_2$ & $16\times21$ & $5\times5$ & - & 12 \\
			H6$\lambda_1\lambda_2$ & $6\times8$ & - & 2 & 14 \\
			\bottomrule
		\end{tabular}
	\end{center}
	\label{table:eff}
\end{table}

The theoretical number of RANSAC iterations is shown in~\ref{fig:ransac_iterations} plotted as the function of outlier ratio in the data. It can be seen that the proposed solver, due to requiring the fewest correspondences, lead to reasonable number of RANSAC iterations even in the most challenging cases.

\section{Conclusions}

In this paper, we propose four new minimal solvers for image stitching, considering different camera configurations and cameras aligned with the gravity direction. 
These configurations include solvers for fixed and varying focal length with or without radial distortion.
The proposed methods are tested on synthetic scenes and on more than $500$k image pairs from publicly available datasets. 
Since we have not found datasets for image stiching with available gravity vector, we captured a new one consisting of 7770 image pairs in total with a smartphone equipped with an IMU sensor. 

{\small
\bibliographystyle{ieee_fullname}
\bibliography{egbib}
}

\end{document}